\documentclass{article}

\usepackage[preprint]{neurips_2025}

\usepackage[utf8]{inputenc}  
\usepackage[T1]{fontenc}   

\usepackage{todonotes}

\usepackage{titletoc}
\usepackage[page, header, toc, page]{appendix} 

\usepackage{hyperref}    
\usepackage{url}        
\usepackage{booktabs}    
\usepackage{amsfonts}  
\usepackage{verbatim}
\usepackage{nicefrac} 
\usepackage{microtype}     
\usepackage{xcolor}    
\usepackage{amsmath,amssymb,amsthm}
  
\usepackage{algorithm}
\usepackage{algpseudocode}
\usepackage{enumitem} 
\usepackage{natbib}
\usepackage{etoolbox}
 
\usepackage{xcolor,graphicx}
\usepackage{colortbl}
\usepackage{tcolorbox}

\usepackage{wrapfig} 
\usepackage{subcaption}
\usepackage{booktabs}
\algnewcommand{\States}[1]{%
\Statex \hspace{-0.5em} \hspace{-\algorithmicindent}\textbf{States:} #1%
} 

\usepackage{multicol}
\usepackage{multirow}

\setlist{nolistsep}
\setlist{leftmargin=*} 

\definecolor{darkgreen}{rgb}{0,0.3,0}
\definecolor{darkblue}{rgb}{0.0,0.0,0.65}
\definecolor{darkred}{rgb}{0.3,0.0,0.0}
\hypersetup{
colorlinks = true,
citecolor  = darkblue,
linkcolor  = darkred,
filecolor  = darkblue,
urlcolor   = darkblue,
}
 
\usepackage{mathtools}
\usepackage{thmtools}

\newcommand{\algname}[1]{\texttt{\textsc{#1}}}
\newcommand{\algcolor}[1]{\textcolor{blue!70!black}{#1}}
\newcommand{\algcomment}[1]{\textcolor{blue!70!black}{\hfill
\small{$\triangleright$\texttt{\hspace{2pt}#1}}}}

\definecolor{lightgray}{gray}{0.6}
\definecolor{DarkGreen}{rgb}{0, 0.5, 0}

\newcommand{\R}{\mathbb{R}}
\newcommand{\E}{\mathbb{E}}

\title{Dion: Distributed Orthonormalized Updates}

\author{
  Kwangjun Ahn\textsuperscript{1} \quad
  Byron Xu\textsuperscript{1} \quad
  Natalie Abreu\textsuperscript{1,}\textsuperscript{2} 
  \quad Ying Fan\textsuperscript{1} \AND
  Gagik Magakyan\textsuperscript{1,}\textsuperscript{3} \quad   Pratyusha Sharma\textsuperscript{1}\quad  Zheng Zhan\textsuperscript{1}   \quad  John Langford\textsuperscript{1} \\\\
  \textsuperscript{1}Microsoft Research, AI Frontiers \\
  \textsuperscript{2}Harvard University 
  \textsuperscript{3}MIT \\
 Correspondence: \texttt{\{kwangjunahn, byronxu\}@microsoft.com}
}

\begin{document}

\maketitle


\begin{abstract}
Orthonormalized updates accelerate training, improve stability, and enable robust hyperparameter transfer, but existing methods like Muon rely on dense matrix operations that clash with sharded weights in large-scale LLM training, causing high compute and communication cost.
We introduce \textbf{Dion} (Distributed Orthonormalization), a scalable and efficient update rule that replaces Newton–Schulz iteration with amortized power iteration on a momentum buffer, avoiding full-matrix reconstruction and integrating cleanly with weight sharding. The rank-fraction parameter with error feedback enables low-rank updates that balance quality with significant cost savings.
On language models from 160M to 3B parameters, Dion retains the benefits of orthonormalized updates, while markedly reducing wall-clock time at scale, making it a practical optimizer for next-generation foundation models.  
\begin{align*}
\boxed{\text{Code is available at: \url{https://github.com/microsoft/dion/}}}    
\end{align*}
\end{abstract}

\begin{figure}[h]
    \centering
\includegraphics[width=\linewidth]{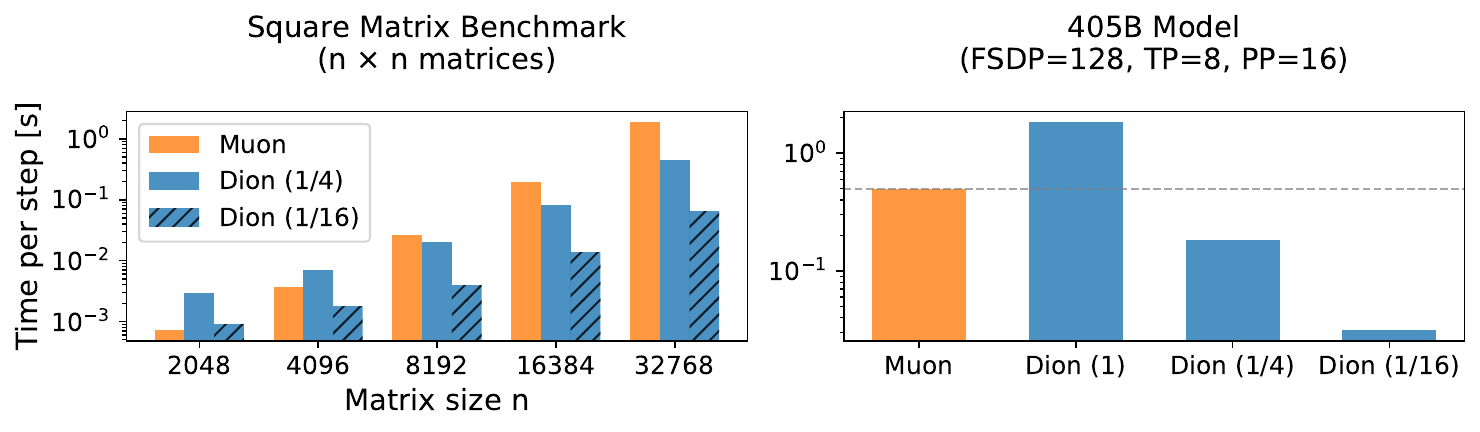}
\caption{\footnotesize Wall-clock time per training step comparison of Muon and Dion optimizers on H100 GPUs (simulated on a single GPU without communication costs). 
    (\textbf{Left}) Benchmark across matrix parameters of sizes 2048$^2$-32768$^2$, showing low-rank Dion achieve faster orthonormal updates at larger scales. (\textbf{Right}) Time per step on Llama-3 405B model~\citep{grattafiori2024llama}, a challenging configuration for Muon \citep{essential2025muon}.  
     }
    \label{fig:bench}
\end{figure}


\newcommand{\AllGather}{\mathbf{AllGather}}
\newcommand{\AllReduce}{\mathbf{AllReduce}}
\newcommand{\AllToAll}{\mathbf{AllToAll}}
\newcommand{\ReduceScatter}{\mathbf{ReduceScatter}}
\newcommand{\red}[1]{\textcolor{red}{#1}}
\newcommand{\blue}[1]{\textcolor{blue}{#1}}
\newcommand{\comsync}{{\texttt{compressed}}}
\newcommand{\true}{\red{\texttt{True}}}
\newcommand{\false}{\blue
{\texttt{False}}}

\newcommand{\btrue}{ {\texttt{True}}}
\newcommand{\bfalse}{ {\texttt{False}}}
 
\newcommand{\offload}{\texttt{cpu}}
\newcommand{\future}{\mathsf{Future}}

\section{Introduction}

Training state-of-the-art AI models consumes millions of GPU-hours, making improvements to the optimizer's \emph{update rule} critical for reducing costs.  In this work, we propose a new update rule for \emph{matrix-valued} parameters, which constitute the bulk of modern neural network weights.

\textbf{Orthonormalized update rules}, exemplified by the Muon optimizer~\citep{jordan2024muon}, has recently shown strong promise. By conditioning weight updates to induce consistent changes in hidden states~\citep{bernstein2025deriving}, they deliver faster convergence, improved training stability, and robust hyperparameter transfer across scales~\citep{bernstein2024modular,large2024scalable,pethick2025training}. Large-scale studies report that Muon nearly doubles the efficiency of AdamW~\citep{liu2025muon}, scales well with batch size~\citep{shah2025practical}, and has been deployed in foundation models such as Kimi-K2~\citep{kimi2025k2} and GLM-4.5~\citep{zeng2025glm}.  

Yet scaling orthonormalized updates exposes a critical systems bottleneck. Muon’s reliance on Newton–Schulz iteration~\citep{bernstein2024modular,jordan2024muon} requires dense matrix multiplications that conflict with how large-scale LLM training shards parameters across devices.   For instance, \citep{essential2025muon} reports that Muon’s overhead can become either compute- or communication-bound when training Llama-3 405B.
This motivates a central question:  
\begin{center}
\emph{Can we devise a compute- and communication-efficient orthonormalized update rule?}     
\end{center}
We introduce \textbf{Dion} (Distributed Orthonormalization), a  scalable update rule that rethinks the underlying linear algebra. Instead of applying Newton–Schulz to the full update matrix, Dion performs \emph{amortized power iteration} on a slowly evolving momentum buffer, avoiding full-matrix reconstruction and integrating seamlessly with weight sharding. Inspired by PowerSGD~\citep{vogels2019powersgd}, Dion decouples the momentum buffer to enable compressed data-parallel synchronization while preserving equivalence to the centralized formulation.  

A key scalability lever is the \emph{rank-fraction} parameter, which allows low-rank updates that significantly reduce compute and communication costs. To enhance update quality, Dion uses \emph{error feedback}, adding the low-rank approximation error in the momentum buffer, allowing persistent signals to accumulate and be expressed in future steps.  Empirically, we observe that the rank required for strong performance grows slowly with model size, and larger models are more robust to rank reduction.  

We evaluate Dion on language models from 160M to 3B parameters, analyzing scaling with both model and batch size. Across all settings, Dion preserves the advantages of orthonormalized updates, while benchmarks (\autoref{fig:bench}) show it delivers substantially lower wall-clock time at the largest scales. Together, these results indicate that Dion retains Muon’s optimization benefits while overcoming its system bottlenecks, making it a practical optimizer for training next-generation foundation models.

\subsection{Why Dion? Advantages over Muon}

Given Muon's success in large-scale AI training~\citep{kimi2025k2,zeng2025glm}, one may ask: what does Dion add? Its key advantages are:

\textbf{Preserved benefits.} Being an orthonormalized update, Dion retains all of Muon’s practical strengths—fast convergence, tolerance of large batch sizes, robust hyperparameter transfer—as verified in \autoref{sec:exp}.

\textbf{Enhanced scalability.} Dion introduces a new lever—the rank-fraction parameter—which trades a small amount of update quality for substantial efficiency gains. As shown in \autoref{fig:bench}, low-rank variants remain practical even at very large matrix sizes, making Dion viable at extreme scales such as Llama-3 405B, where Muon encounters bottlenecks~\citep{essential2025muon}.

\textbf{Seamless weight sharding.} Dion supports efficient implementations under both 1D sharding (\autoref{alg:1d-dion}) and 2D sharding (\autoref{alg:2d-dion}), ensuring compatibility with distributed training.

\textbf{Algorithmic flexibility.} By rethinking the underlying linear algebra—replacing Newton–Schulz iterations with amortized power iteration—Dion enables new forms of optimization, as demonstrated in \autoref{sec:flexible}.  
For instance, combining lazy updates with CPU offloading yields orthonormalized updates with only minimal wall-clock overhead (\autoref{sec:lazy}).

\section{Notation Conventions}
\label{sec:scalingbook-notation}

\begin{algorithm}[t]
\caption{A single step of Dion optimizer on a (sharded) matrix  parameter $X\in \R^{m\times n}$.}
\label{alg:un-dion}
\begin{algorithmic}[1]
\Statex \textbf{Hyperparams:} Rank factor $r$, learning rate $\eta$.

\Statex \textbf{Optimizer states:} momentum $M\in \R^{m\times n}$, right vectors $V \in \R^{n\times r}$.

\State Receive gradient $G\in\R^{m\times n}$
 
\State \(O,\,M_{\mathrm{new}},\,V_{\mathrm{new}} \gets \begin{cases}
\algname{Dion}^{0\mathrm{D}}(G,M,V) & (\text{\autoref{alg:un-dion}})   \quad \text{if } X \text{ is unsharded},  \\
\algname{Dion}^{1\mathrm{D}}(G,M,V)  & (\text{\autoref{alg:1d-dion}})   \quad \text{if }X \text{ is 1D-sharded}, \\ 
\algname{Dion}^{2\mathrm{D}}(G,M,V) & (\text{\autoref{alg:2d-dion}}) \quad \text{if } X \text{ is 2D-sharded},\\
\begin{cases}
    \algname{Lazy-Dion}_k(G,M,V)\\
    \algname{CPU-Dion}(G,M,V)
\end{cases} & (\text{\autoref{alg:lazy-dion}}) \quad \text{for faster variants.}
\end{cases}\)
\State \(X_{\mathrm{new}} \gets X - \eta\, O\)
\end{algorithmic}
\end{algorithm}

We adopt the sharded array and communication notation of
``How To Scale Your Model''~\citep[Ch.~3]{scaling-book}, which expresses
parallelism and sharding directly in tensor indices.
Below we give a concise overview of the notation, while referring readers
to the reference for full details.

A device mesh is written
\(\text{Mesh}(\{\texttt{X}:a, \texttt{Y}:b, \texttt{Z}:c\})\), where
each symbol denotes a named mesh axis. An array \(A[I,J,\ldots]\) is annotated as
\(A[I_{\texttt{X}}, J_{\texttt{Y}}, \ldots]\) to indicate that dimension
\(I\) is sharded across mesh axis \(\texttt{X}\), \(J\) across
\(\texttt{Y}\), etc. Dimensions without subscripts are replicated.
For example, \(X[I,J_{\texttt{X}}]\) denotes a matrix of shape
\((|I|,|J|)\) column-sharded across \(\texttt{X}\).
Low-rank factorizations introduce a dedicated rank axis \(R\) with size
\(|R|=r\).  

Matrix multiplications are written with explicit contracted indices:
\[
C[I,K] \gets A[I,J] \cdot_{J} B[J,K],
\]
which makes clear which logical axis is eliminated. 
If \(J\) is sharded across \(\texttt{X}\) (e.g.\ \(A[I,J_{\texttt{X}}]\) and
\(B[J_{\texttt{X}},R]\)), the local computation produces a \emph{partial} result
along \(\texttt{X}\), written with an unreduced tag: $C[I,K]\{U_{\texttt{X}}\}$. In other words, 
\(\{U_{\texttt{X}}\}\) means “this quantity still needs a reduction over the
\(\texttt{X}\) mesh axis.”  
Applying a collective that reduces \(\texttt{X}\) clears the tag; for instance, 
\( \AllReduce_{\texttt{X}}\!\big(C[I,K]\{U_{\texttt{X}}\}\big).
\)
 
Collectives are written as explicit shape transforms:
\begin{itemize}[topsep=0pt,itemsep=0pt]
\item \(\AllGather_{\texttt{X}}\): removes a subscript \(_{\texttt{X}}\).
\item \(\ReduceScatter_{\texttt{X},K}\): reduces along \(\texttt{X}\) and
  shards the result across logical axis \(K\).
\item \(\AllReduce_{\texttt{X}}\): equivalent to
  \(\ReduceScatter+\AllGather\). 
\item \(\AllToAll_{\texttt{X},K}\): moves a subscript \(_{\texttt{X}}\)
  from one logical axis to another \(K\).
\end{itemize}
We use the following axis notation throughout the paper:
\[
\texttt{X} = \text{Sharding Axis 1}, \quad
\texttt{Y} = \text{Sharding Axis 2}, \quad
\texttt{Z} = \text{Replicated Data Parallel}.
\]
For example, standard DDP corresponds to replication along \texttt{Z} with no sharding on \texttt{X} or \texttt{Y}.  
When either sharding axis is combined with \texttt{Z}, the configuration yields hybrid sharding—FSDP within nodes together with replicated data parallel across nodes~\citep{zhao2023pytorch}.


\section{Dion without Weight Sharding}
\label{sec:dion}

For intuition, we begin by considering a simplified scenario without weight sharding. We consider optimizing a parameter matrix
$X$ with dimensions $|I|\times |J|$, where $|I|$ and $|J|$ are the ``out'' and ``in'' dimensions. Let $G$ denote the current gradient, and $M$ the momentum buffer.

\newcommand{\dpsynctag}{\texttt{\small compressed DP-sync}}
\newcommand{\DPsync}[1]{%
  \begingroup
    \color{red}#1%
  \endgroup\hfill\(\triangleright\) \red{\dpsynctag}%
} 
\newcommand{\DPsyncLegend}{\(\triangleright\) \red{\dpsynctag}\;=\; compressed data-parallel synchronization over mesh axis \texttt{Z}.}
 
\begin{algorithm}[t]
\caption{Unsharded Dion update rule  on $[I,J]$ (where $R$ denotes rank axis $|R|=r$).}
\label{alg:un-dion}
\begin{algorithmic}[1] 
   
\Function{$\algname{Dion}^{0\mathrm{D}}$}{$\mbox{gradient } G[I,J], \mbox{momentum } M[I,J], \mbox{semi-orthonormal } V[J,R]$}
\State \(M[I,J] \gets M[I,J] + G[I,J]\) \algcomment{accumulate gradient}

\State $U[I,R], W[J,R] \gets \algname{PowerIter1}(M, V)$ \algcomment{rank-$r$ approximate $M \approx UW^\top$}

\State \(M_{\mathrm{new}}[I,J] \gets M[I,J] - \beta\, U[I,R]\cdot_{R} W^\top[R,J]\) \algcomment{$\beta=0.05$ error feedback; see \eqref{exp:error-feed}}
\State \(V_{\mathrm{new}}[J,R] \gets \texttt{ColNorm}\big(W[J,R]\big)\)  \algcomment{column normalization}
\State \(O[I,J] \gets U[I,R]\cdot_{R}V_{\mathrm{new}}^\top[R,J]\)  \algcomment{orthonormal update}
\State \textbf{return} \(\sqrt{|I|/|J|}\;O[I,J] ,\; M_{\mathrm{new}}[I,J],\; V_{\mathrm{new}}[J,R]\)
\EndFunction
\end{algorithmic}

\vspace{-0.8em}
\rule{\textwidth}{0.4pt}
\vspace{-1.3em}

\begin{algorithmic}[1]
\Function{$\texttt{PowerIter1}$}{$\mbox{momentum } M[I,J], \mbox{ semi-orthonormal } V[J,R]$}  
\State \(P[I,R] \gets M[I,J]\cdot_{J} V[J,R]\)
  \State \DPsync{\(P[I,R] \gets \AllReduce_{\texttt{Z}}\big(P[I,R]\{U_{\texttt{Z}}\}\big)\)}  
 
\State \(U[I,R] \gets \algname{Orthonormalize}(P[I,R])\) \algcomment{see \autoref{sec:orthonormalize}}
\State \(W[J,R] \gets M^\top[J,I]\cdot_{I} U[I,R]\)
\State  \DPsync{\(W[J,R] \gets \AllReduce_{\texttt{Z}}\big(W[J,R]\{U_{\texttt{Z}}\}\big)\)}  
 
\State \textbf{return} $U[I,R], W[J,R]$ \algcomment{$U$ is orthonormal}
\EndFunction
\end{algorithmic}

\vspace{0.3em}\noindent\DPsyncLegend
\end{algorithm}

\subsection{Design Intuition Behind Dion}

The Muon optimizer~\citep{jordan2024muon} introduces an effective update rule for matrix parameters based on approximate zeroth-power orthonormalization. Muon maintains a momentum matrix that accumulates gradients over time: $M \gets (1-\beta) M + G$, for $\beta=0.05$ 
and then applies an update based on the zeroth power of this matrix. Specifically, it approximates
$U V^\top$ where $M = U \Sigma V^\top$ 
is the SVD of the momentum matrix.
To approximate this zeroth-power orthonormalization efficiently, \citet{jordan2024muon} employ Newton-Schulz iterations. However, Newton-Schulz iteration requires computing full matrix-matrix products, which faces challenges in distributed training.

To address the limitations of Muon, we rethink the underlying linear algebra: for rank $r < |I|, |J|$, we efficiently compute a rank-$r$ approximation of the momentum matrix using ``amortized'' power iteration, followed by column normalization to ensure the update is orthonormal. Using a technique from PowerSGD \citep{vogels2019powersgd}, we warm-start power iteration with the result from the previous optimizer step, so that a single step of power iteration suffices to produce an accurate low-rank approximation.
We empirically validate the effectiveness of this approach through an ablation study comparing it with a low-rank approximation using SVD, presented in \autoref{sec:ablate}.

Because a low-rank update cannot capture all the information in its input, we apply an \textbf{error feedback} mechanism to compensate. 
Unlike PowerSGD's error feedback, we reuse the momentum buffer itself and require no additional optimizer state memory. 
In \autoref{sec:ablate}, we present an ablation study demonstrating the importance of error feedback.

Specifically, at each step, we first form the buffer $M \gets M + G$ and use power iteration to compute a rank-$r$ approximation
\[ 
U[I,R] \cdot_R W^\top [R,J] \approx M[I,J], \quad \text{where}\ \text{$|R|=r$.}
\]  
We then incorporate the error of approximation into the next momentum update:
\begin{align}\label{exp:error-feed}
M_{\text{new}} \gets (1-\beta) M  + \beta \Delta  \quad \text{where} \quad \Delta  \coloneqq  M  -  U W^\top.
\end{align}
This way, information not captured by the low-rank approximation may propagate across iterations.
This is equivalently written as (as presented in \autoref{alg:un-dion}): $M_{\text{new}} \gets M - \beta U W^\top$.

Lastly, we apply a scaling factor of $\sqrt{|I|/|J|}$ to the orthonormalized update to improve learning rate transferability across different model sizes \citep{bernstein2024modular,pethick2025training}. We elaborate on this scaling in \autoref{sec:scalar}.

\subsection{\texttt{Orthonormalize} in Power Iteration}
 \label{sec:orthonormalize}
The \texttt{Orthonormalize} routine in \texttt{PowerIter1} (\autoref{alg:un-dion}) can be implemented using a standard QR decomposition or faster variants. 
For two-way sharded weights, distributed randomized Cholesky QR (RCQR; \autoref{alg:dist-orth}) is preferable, as it avoids reconstructing the full matrix on a single device (\autoref{sec:2d-dion}). 
Further, in \autoref{sec:cholesky} we examine replacing QR/RCQR with plain Cholesky, which provides additional speedups for power iteration.

\subsection{Compressed Data-Parallel Gradient Sync}
\label{sec:compressed-sync}

\begin{table}[h]
\centering\small
\begin{tabular}{@{}lccc@{}}
\toprule
\textbf{Mode} & \textbf{DP Sync} & \textbf{DP I/O} & \textbf{Momentum} \\
\midrule
  Usual DP-sync & Full gradient & $mn$ & Synchronized \\
 \red{\dpsynctag} & Low-rank states & $(m+n)r$ & Decoupled \\
\bottomrule
\end{tabular}
\vspace{6pt}
\caption{Data-parallel synchronization modes. ``\textcolor{black}{\dpsynctag}'' reduces DP communication I/O from $\mathcal{O}(mn)$ to $\mathcal{O}((m+n)r)$ for an $m \times n$ weight matrix.}
\label{tab:compressed-sync}
\end{table}

Dion can bypass conventional full-gradient all-reduce by synchronizing only low-rank matrices instead of complete gradients. This technique, inspired by PowerSGD \citep{vogels2019powersgd}, reduces communication volume while producing identical results. Depending on the rank fraction, communication requirements can be reduced by orders of magnitude.

This compressed synchronization works with any replicated data-parallel axis, including standard DDP and hybrid sharding setups \citep{zhao2023pytorch}; see \autoref{sec:hybrid} for hybrid sharding details.  With ``\dpsynctag,'' the optimizer employs \emph{decoupled momentum} similar to DeMo \citep{peng2024decoupled}: internal momentum states $M$ diverge across data-parallel processes while weight updates remain identical. Crucially, both synchronization methods produce mathematically identical weight updates—only the communication pattern differs, not the optimization trajectory.

\subsection{Non-Matrix Parameters and Learning Rate Compatibility} 
\label{sec:scalar} 
\begin{table}[h] 
\centering\small
    \begin{tabular}{@{}cccc@{}} 
    \toprule
    Parameter & Shape & Norm & Scaling factor \\
    \midrule
    Weight & $d_{\text{out}} \times d_{\text{in}}$ & \texttt{Spectral} & $\sqrt{d_{\text{out}} / d_{\text{in}}}$ \\
    Bias & $d_{\text{out}} \times 1$ & \texttt{RMS} & 1 \\
    Embedding & $d_{\text{out}} \times 1$ & \texttt{RMS} & 1 \\
    LM Head & $1 \times d_{\text{in}}$ & \texttt{RMS} & $1 / \sqrt{d_{\text{in}}}$ \\
    Normalization & $1 \times 1$ & \texttt{RMS} & 1 \\
    \bottomrule
\end{tabular}
\vspace{3pt}
\caption{ \footnotesize Normalization and scaling factors for various parameter types. For learning rate transfer, the parameter update should be unit-normed in the specified norm, and the base learning rate should be multiplied by the specified scaling factor. Refer to \autoref{sec:non-matrix} for a more detailed explanation.}
\label{tab:scaling-factors}
\end{table}

Dion's orthonormal update applies to matrix-shaped weights, while other parameters (biases, embeddings, normalization factors) use element-wise optimizers like AdamW or Lion, following the approach in \cite{jordan2024muon} and \cite{liu2025muon}.
To handle simultaneous use of two optimizers, we scale a \textbf{shared base learning rate} for each parameter type and shape. This leads to Dion's hyperparameter transfer properties across model sizes (see \autoref{sec:transfer}).

Since these optimizers produce unit-normed updates—Dion/Muon via orthonormalization (spectral norm $= 1$), Lion via sign function (RMS norm $= 1$), and Adam with approximately constant RMS norm \citep{liu2025muon}—we follow \cite{yang2023spectral} and scale the base learning rate per-parameter to maintain roughly constant input-to-output mappings. Scaling factors appear in \autoref{tab:scaling-factors}, with detailed derivation in \autoref{sec:non-matrix}.

\subsection{Computational Complexity (Unsharded)}
\label{sec:comp-unshard}

For a single unsharded parameter matrix $X \in \R^{m\times n}$ at rank $r$, the detailed FLOP analysis in \autoref{details:comp} shows that Dion requires
\begin{align} \label{flop:un-dion}
    {8mnr} 
\;+\; {6.5\,mr^2} 
\;+\;{2.17\,r^3} 
\;+\;O(mn) \quad \text{FLOPs}.
\end{align}
By contrast, Muon’s five Newton–Schulz iterations cost
\(20mn^2 + 10n^3 + O(mn)\) FLOPs \citep{jordan2024muon, liu2025muon}.
Even in the full-rank setting ($r=n$), Dion’s complexity is $14.5\,mn^2 + 2.17\,n^3$,
which is strictly lower than Muon’s. Thus, Dion achieves lower asymptotic cost while retaining the same orthonormalization guarantee.

\section{Dion with Weight Sharding}
\label{sec:dist-dion}

In distributed training, weights are typically sharded across  devices. We present two distributed implementations: 1D-Sharded Dion (\autoref{alg:1d-dion})  and 2D-Sharded Dion (\autoref{alg:2d-dion}).

\subsection{1D-Sharded Dion}
\label{sec:1d-dion}

\begin{algorithm}[t]
\caption{1D-Sharded Dion update rule on $[L,I,J_{\texttt{X}}]$ (\algcolor{where $|L|=|\texttt{X}|$})}
\label{alg:1d-dion}
\begin{algorithmic}[1]

\Statex  \red{\textit{Note: Compressed DP sync along $\texttt{Z}$ is omitted for brevity (it follows \autoref{alg:un-dion}).}}

\Statex \textbf{Optimizer states:} \(\mbox{momentum }M[L,I,J_{\texttt{X}}],\,\mbox{semi-orthonormal }V[L,J_{\texttt{X}},R]\).

 \Function{$\algname{Dion}^{1\mathrm{D}}$}{$\mbox{gradient } G[L,I,J_{\texttt{X}}],\, M[L,I,J_{\texttt{X}}],\, V[L,J_{\texttt{X}},R]$}
 
\State \(M[L,I,J_{\texttt{X}}] \gets M[L,I ,J_{\texttt{X}}] + G[L,I,J_{\texttt{X}}]\)
\State \(P[L,I,R]\{U_{\texttt{X}}\} \gets M[L,I,J_{\texttt{X}}]\cdot_{J} V[L,J_{\texttt{X}} ,R]\)
 
\State \( P[L_{\texttt{X}},I,R] \gets \ReduceScatter_{\texttt{X},L}\big(P[L,I,R]\{U_{\texttt{X}}\}\big)\)  
\State \(U[L_{\texttt{X}},I,R] \gets \algname{Orthonormalize}(P[L_{\texttt{X}},I,R])\)
 
\State \(U[L,I,R] \gets \AllGather_{\texttt{X}}\big(U[L_{\texttt{X}},I,R]\big)\)

\State \(W[L,J_{\texttt{X}},R]  \gets M^\top[L,J_{\texttt{X}},I]\cdot_{I} U[L,I ,R]\) 
 
\State \(M_{\mathrm{new}}[L,I,J_{\texttt{X}}] \gets M[L,I, J_{\texttt{X}}] - \beta\, U[L,I,R]\cdot_{R}W^\top[L,R,J_{\texttt{X}}]\)
\State \(V_{\mathrm{new}}[L,J_{\texttt{X}},R] \gets \algname{ColNorm}\big(W[L,J_{\texttt{X}},R]\big)\)  \algcomment{$\AllReduce^{\text{sum}}_{\texttt{X}}$ of columnn norms}
\State \(O[L,I,J_{\texttt{X}}] \gets U[L,I ,R]\cdot_{R}V_{\mathrm{new}}^\top[L,R,J_{\texttt{X}}]\)
\State \textbf{return} \(\sqrt{|I|/|J|}\; O[L,I ,J_{\texttt{X}}],\; M_{\mathrm{new}}[L,I,J_{\texttt{X}}] ,\; V_{\mathrm{new}}[L,J_{\texttt{X}},R]\)
\EndFunction
\end{algorithmic}
\end{algorithm}

The key idea is to adapt the power iteration procedure from \autoref{alg:un-dion} to sharded tensors while preserving mathematical equivalence.  
We consider a weight matrix $X$ sharded across $|\texttt{X}|$ devices along the input dimension $J$, with the batch dimension $L$ grouping matrices of identical shape for efficient batched computation.  
Each device stores a shard $X[L,I,J_{\texttt{X}}]$, and the optimizer states $M$ and $V$ are sharded accordingly.  
For simplicity, we assume $|L|=|\texttt{X}|$ after suitable batching or padding.  

\textbf{1D-Sharded Dion} (\autoref{alg:1d-dion}) operates as follows.  
Gradients are first accumulated into the local momentum shard $M$, which is multiplied with the local factor $V$ to form partial products.  
These are aggregated via $\ReduceScatter$ and then scattered across the $L$ dimension, yielding the layer-sharded matrix $P[L_{\texttt{X}},I,R]$, which is orthonormalized to produce $U[L_{\texttt{X}},I,R]$.  
An $\AllGather$ across $\texttt{X}$ then reconstructs $U[L,I,R]$.  
The remaining steps mirror \autoref{alg:un-dion}, with one difference: column normalization of $W$ requires an additional $\AllReduce^{\text{sum}}_{\texttt{X}}$ to compute global column norms.  
This operation is lightweight, as it synchronizes only $O(r)$ scalars.

 \subsection{Dion for Hybrid Sharding}
\label{sec:hybrid}

Hybrid sharding combines fully-sharded data parallel (FSDP) within nodes and data parallel (DP) across nodes, providing a tunable trade-off between memory efficiency and throughput \citep{zhao2023pytorch}. This approach is motivated by both model scaling requirements and system topology constraints, where there is fast intra-node bandwidth but limited inter-node connectivity. 

However, DP replicas spanning multiple nodes still require gradient synchronization over limited inter-node bandwidth, making traditional all-reduce operations costly.
Dion's compressed gradient synchronization (\autoref{sec:compressed-sync}) is particularly beneficial in this scenario. Instead of synchronizing full gradients, we only communicate compressed low-rank states, reducing inter-node traffic.

\subsection{2D-Sharded Dion}
\label{sec:2d-dion}

 At very large scale, training often employs two-way sharding (e.g.\ \texttt{X}$\times$\texttt{Y}) to maximize hardware utilization. 
We extend Dion to this setting, where weights are sharded on both logical axes:
$X[L,I_{\texttt{Y}},J_{\texttt{X}}]$.
As in the 1D-sharded case, reductions for $P = M \cdot V$ happen across $\texttt{X}$, 
while reductions for $W = M^\top \cdot U$ happen across $\texttt{Y}$, preserving locality of both multiplies.

The pseudocode supports two batching regimes:   
\begin{itemize}
    \item (\emph{i}) \emph{small batch}, $|L| = |\texttt{Y}|$, where partial $P$ is summed with $\AllReduce_{\texttt{X}}^{\text{sum}}$;  
    \item (\emph{ii}) \emph{large batch}, $|L| = |\texttt{X}|\cdot |\texttt{Y}|$, where partial $P$ is first reduced via $\ReduceScatter_{\texttt{X},L}$ and later reassembled with $\AllGather_{\texttt{X}}$. 
\end{itemize}
Both produce the same \(U[L,I_{\texttt{Y}},R]\), trading memory for bandwidth.

\textbf{Distributed orthonormalizaton.}
A full QR on \(P\) requires materializing the unsharded \(I\) dimension and thus breaks 2D sharding. 
Instead, \autoref{alg:2d-dion} computes an orthonormal basis \(U\) for the columns of \(P\) using a two–stage randomized Cholesky-QR that preserves \(\texttt{Y}\)-sharding and communicates only \(r\times r\) matrices \citep{fan2021novel,epperly2024neat}. 
Concretely (see \autoref{alg:dist-orth}):
\begin{enumerate}[leftmargin=*,label=(\roman*)]
    \item \textbf{Sketch \& factor a small matrix.} Draw \(S[\tilde R,I_{\texttt{Y}}]\) with \(|\tilde R|\!\approx\!1.25r\) \citep{melnichenko} and form \(\tilde P=S\cdot_I P\) locally. 
    Apply \(\ReduceScatter_{\texttt{Y},L}\) to \(\tilde P\) and compute a thin QR on the small \((|\tilde R|\!\times\! r)\) sketch, retaining only the triangular \(R_1\); then \(\AllGather_{\texttt{Y}}\) \(R_1\).
    \item \textbf{Pre-whiten.} Solve \(B = P\,R_1^{-1}\) (triangular solve), which balances the column scales of \(P\).
    \item \textbf{Gram \& Cholesky on \(r\times r\).} Form \(G = B^\top\!\cdot_I B\) locally, apply \(\ReduceScatter_{\texttt{Y},L}\) to sum \(G\) across \(\texttt{Y}\), and compute \(R_2=\texttt{Cholesky}(G)\); then \(\AllGather_{\texttt{Y}}\) \(R_2\).
    \item \textbf{Finalize.} Solve \(U = B\,R_2^{-1}\), yielding \(U\) with orthonormal columns.
\end{enumerate}
This procedure performs two  (\(\ReduceScatter_{\texttt{Y}}+\AllGather_{\texttt{Y}}\)), each moving only \(\mathcal{O}(r^2)\) scalars. 
As an alternative, one can temporarily re-shard by \(I\) using two $\AllToAll{}_{\texttt{Y}}$ operations (to gather $P$ for a local QR and to scatter $U$ back), moving \(\mathcal{O}(\nicefrac{r\cdot |I|}{|\texttt{Y}|} )\) elements per device. 
This all-to-all route can be preferable when the fabric favors $\AllToAll{}$ and \(\nicefrac{r\cdot |I|}{|\texttt{Y}|}\) is small relative to $r^2$.

\textbf{Other variants.}  
With two sharding axes \(\texttt{X},\texttt{Y}\), there are two independent binary choices, yielding four symmetric variants:  
(i) orthonormalization (and its reductions) can be placed on either \(\texttt{X}\) or \(\texttt{Y}\);  
(ii) power iteration can be run on \(M\) (as in \autoref{alg:2d-dion}) or on \(M^\top\) (the transposed form).  
These variants are equivalent up to axis renaming and transposition.  
In practice, one should (a) shard \(\texttt{X}\) and \(\texttt{Y}\) along different logical dimensions, (b) place orthonormalization on the axis with faster collectives, and (c) choose the standard or transposed form to match whether the parameter is column- or row-sharded.

 \begin{algorithm}[t]
\caption{2D-Sharded Dion update rule on $[L,I_{\texttt{Y}},J_{\texttt{X}}]$ (\algcolor{where $|L|=|\texttt{Y}|$ or $|L|=|\texttt{X}|\cdot|\texttt{Y}|$})}
\label{alg:2d-dion}
\begin{algorithmic}[1]

\Statex  \red{\textit{Note: Compressed DP sync along $\texttt{Z}$ is omitted for brevity (it follows \autoref{alg:un-dion}).}}

\Statex \textbf{Optimizer states:} \(\mbox{momentum }M[L,I_{\texttt{Y}},J_{\texttt{X}}],\,\mbox{semi-orthonormal }V[L,J_{\texttt{X}},R]\).

\Function{$\algname{Dion}^{2\mathrm{D}}$}{$\mbox{gradient }G[L,I_{\texttt{Y}},J_{\texttt{X}}],\,M[L,I_{\texttt{Y}},J_{\texttt{X}}],\,V[L,J_{\texttt{X}},R]$}

\State \(M[L,I_{\texttt{Y}},J_{\texttt{X}}] \gets M[L,I_{\texttt{Y}},J_{\texttt{X}}] + G[L,I_{\texttt{Y}},J_{\texttt{X}}]\)
\State \(P[L,I_{\texttt{Y}},R]\{U_{\texttt{X}}\} \gets M[L,I_{\texttt{Y}},J_{\texttt{X}}]\cdot_{J} V[L,J_{\texttt{X}},R]\)

\If{\algcolor{$|L|=|\texttt{Y}|$}} 
  \State \(P[L,I_{\texttt{Y}},R] \gets \AllReduce_{\texttt{X}}^{\mathrm{sum}}\!\big(P[L,I_{\texttt{Y}},R]\{U_{\texttt{X}}\}\big)\)
  \State \(U[L,I_{\texttt{Y}},R] \gets \algname{Distributed-Orthonormalize}_{\texttt{Y}}\!\big(P[L,I_{\texttt{Y}},R]\big)\)
\Else \quad{\algcolor{($\triangleright$ $|L|=|\texttt{X}|\cdot|\texttt{Y}|$)}}
  \State \(P[L_{\texttt{X}},I_{\texttt{Y}},R] \gets \ReduceScatter_{\texttt{X},L}\!\big(P[L,I_{\texttt{Y}},R]\{U_{\texttt{X}}\}\big)\)
  \State \(U[L_{\texttt{X}},I_{\texttt{Y}},R] \gets \algname{Distributed-Orthonormalize}_{\texttt{Y}}\!\big(P[L_{\texttt{X}},I_{\texttt{Y}},R]\big)\)
  \State \(U[L,I_{\texttt{Y}},R] \gets \AllGather_{\texttt{X}}\!\big(U[L_{\texttt{X}},I_{\texttt{Y}},R]\big)\)
\EndIf
 
\State \(W[L,J_{\texttt{X}},R] \gets \AllReduce_{\texttt{Y}}^{\mathrm{sum}}\!\big(M^\top[L,J_{\texttt{X}},I_{\texttt{Y}}]\cdot_{I} U[L,I_{\texttt{Y}},R]\big)\)

\State \(M_{\mathrm{new}}[L,I_{\texttt{Y}},J_{\texttt{X}}] \gets M[L,I_{\texttt{Y}},J_{\texttt{X}}] - \beta\, U[L,I_{\texttt{Y}},R]\cdot_{R}W^\top[L,R,J_{\texttt{X}}]\)

\State \(V_{\mathrm{new}}[L,J_{\texttt{X}},R] \gets \algname{ColNorm}\big(W[L,J_{\texttt{X}},R]\big)\) \algcomment{$\AllReduce^{\text{sum}}_{\texttt{X}}$ of column norms}
\State \(O[L,I_{\texttt{Y}},J_{\texttt{X}}] \gets U[L,I_{\texttt{Y}},R]\cdot_{R}V_{\mathrm{new}}^\top[L,R,J_{\texttt{X}}]\)
\State \textbf{return} \(\sqrt{|I|/|J|}\; O[L,I_{\texttt{Y}},J_{\texttt{X}}],\; M_{\mathrm{new}}[L,I_{\texttt{Y}},J_{\texttt{X}}],\; V_{\mathrm{new}}[L,J_{\texttt{X}},R]\)

\EndFunction
\end{algorithmic}
 \end{algorithm}

\begin{algorithm}[t]
\caption{Distributed Orthonormalization (\(\texttt{Y}\)-sharded)}
\label{alg:dist-orth}
\begin{algorithmic}[1]
\Function{$\algname{Distributed-Orthonormalize}_{\texttt{Y}}$}{$P[L',I_{\texttt{Y}},R]$}
\State Sample sketching matrix $S[\tilde{R}, I_{\texttt{Y}}] \sim \mathcal{N}(0,1/\sqrt{1.25r})$, with $|\tilde{R}|=1.25r$
\State $\_\_,R_1[L'_{\texttt{Y}},R,R] \gets \texttt{QR}\!\left(\ReduceScatter_{\texttt{Y},L}\big(S\cdot_{I} P\big)\right)$ \algcomment{factor on small sketch}
\State $R_1[L',R,R] \gets \AllGather_{\texttt{Y}}(R_1[L'_{\texttt{Y}},R,R])$ \algcomment{broadcast $R_1$ via $\AllGather_{\texttt{Y}}$}
\State $B[L',I_{\texttt{Y}},R] \gets P[L',I_{\texttt{Y}},R]\;R_1^{-1}$ \algcomment{apply triangular solve}
\State $R_2[L'_{\texttt{Y}},R,R] \gets \texttt{Cholesky}\!\left(\ReduceScatter_{\texttt{Y},L}(B^\top \cdot_{I} B)\right)$
\State  $R_2[L',R,R] \gets \AllGather_{\texttt{Y}}(R_2[L'_{\texttt{Y}},R,R])$ \algcomment{broadcast $R_2$ via $\AllGather_{\texttt{Y}}$}
\State $U[L',I_{\texttt{Y}},R] \gets B[L',I_{\texttt{Y}},R]\;R_2^{-1}$ \algcomment{final triangular solve}
\State \textbf{return} $U[L',I_{\texttt{Y}},R]$
\EndFunction
\end{algorithmic}
\end{algorithm}

\subsection{Compute/Communication Complexity}
\label{sec:complexity}

We analyze the cost of 2D-Sharded Dion, noting that 1D-sharded Dion is the special case with $|\texttt{Y}|=1$.
Let $m=|I|$, $n=|J|$, $r=|R|$, and let $a=|\texttt{X}|$, $b=|\texttt{Y}|$ be the shard counts on the two model-parallel axes (with data-parallel replicas along \texttt{Z}).

\textbf{Computation.}
For 2D-sharded Dion, the per-device FLOP count depends on the batching regime:

\begin{itemize}[leftmargin=*]
\item \textbf{Large batch} ($|L|=ab$): all work is evenly distributed across both axes, so each device performs exactly $\tfrac{1}{ab}$ of the unsharded cost in \eqref{flop:un-dion}.
\item \textbf{Small batch} ($|L|=b$): the per-device cost is
\( \tfrac{8mnr}{ab} + \tfrac{6.5\,mr^2}{b} + \tfrac{2.17\,r^3}{b}\). 
Relative to \eqref{flop:un-dion}, the $O(mnr)$ multiplications are distributed across both $a$ and $b$, while the $O(mr^2)$ and $O(r^3)$ terms are only shared across $b$.
\end{itemize}

 \textbf{Communication.}
We omit DP synchronization along \texttt{Z} (see \autoref{sec:compressed-sync}) and focus on model-parallel collectives along \texttt{X} and \texttt{Y}. 

\begin{itemize}[leftmargin=*]
\item Along \texttt{X}, the dominant cost is exchanging $P$ (and later $U$), which scales as $O(mr)$ elements per device, plus a lighter $O(|L|r)$ term from column-norm aggregation.
\item Along \texttt{Y}, the dominant cost is exchanging $W$, which scales as $O(nr)$ elements per device. The distributed orthonormalization adds only $O(r^2)$ micro-collectives.
\end{itemize}

Overall, communication depends on $mr$ and $nr$ (linear in the matrix dimensions and rank), rather than on the full parameter size $mn$. This makes Dion communication-efficient even at extreme scales. Precise per-device volumes for each collective are given in \autoref{app:comm-details}.

\section{Algorithmic Flexibility of Dion}
\label{sec:flexible}

Replacing Newton–Schulz with amortized power iteration gives Dion 
substantial algorithmic flexibility. This shift enables a range of 
variants that reduce computation and communication overhead while 
retaining the benefits of orthonormalized updates. In this section, 
we explore three directions: (i) \emph{lazy} and CPU-overlapped 
orthonormalization for faster high-rank updates, (ii) two-stage designs 
that minimize replicated-DP traffic under severe bandwidth constraints, 
and (iii) adaptive filtering based on effective rank to cut redundant 
update directions.

\subsection{Faster Dion with Lazy Orthonormalization}
\label{sec:lazy}
For higher ranks, the dominant cost in Dion is the orthonormalization step 
(QR, RCQR, or Cholesky-QR) used in power iteration. 
A key observation is that the momentum matrix exhibits 
\textbf{slow spectral drift}: the leading singular directions of $M$ 
evolve only gradually across training steps. 
Recomputing them at every iteration is therefore wasteful, 
and Dion can be accelerated by updating the principal directions more lazily.  \autoref{alg:lazy-dion} illustrates two complementary strategies, and empirical validation is presented in \autoref{sec:fast-exp}.

\textbf{Lazy-Dion.} This variant maintains the right subspace estimate 
$V \in \mathbb{R}^{|J|\times |R|}$ and refreshes it only once every $k$ 
steps using a symmetric power iteration ($\algname{SymPowerIter1}$ 
in \autoref{alg:lazy-dion}). 
Between recomputations, it avoids QR entirely and reuses the cached $V$ 
to form the low-rank update. 
The symmetric formulation greatly improves stability under infrequent updates 
compared to the standard one-sided power iteration ($\algname{PowerIter1}$).

\textbf{CPU-Dion.} An alternative is to overlap orthonormalization 
with model training. Instead of refreshing $V$ synchronously, 
CPU-Dion asynchronously launches $\algname{SymPowerIter1}$ on a CPU copy of $M$, 
while the GPU continues executing forward/backward passes. 
Each iteration uses the most recent available $V$ (possibly lagged by one or 
more steps). When the CPU task finishes, $V$ is updated and a new job is 
spawned. Because CPUs are often underutilized in large-scale training, 
this overlap amortizes the cost of orthonormalization and makes 
high-rank Dion practical even at extreme scales.

\textbf{Complexity.} Between recomputations, the incremental cost reduces 
to three matrix multiplications (GEMMs) plus a column normalization. 
With CPU offloading, the recomputation itself is hidden behind GPU work; 
the only added overhead is transferring $M$ (or its shards) to host memory, 
which can also be overlapped with forward/backward execution.

\begin{algorithm}[t]
\caption{Faster Dion variants update rule on $[I,J]$ (unsharded; sharded follows Dion).}
\label{alg:lazy-dion}
  
\begin{algorithmic}[1]
\Function{$\algname{Lazy-Dion}^{0\mathrm{D}}_{k}$}{$\mbox{gradient } G[I,J],\,\mbox{momentum }M[I,J],\,\mbox{semi-orthonormal }V[J,R]$}
\State $t \gets t+1$;\quad $M \gets M + G$    
\If{$t \bmod k = 0$} \algcomment{every $k$ steps}
     
        \State $V \gets \algname{SymPowerIter1}(M, V)$  \algcomment{refresh $V$ with power iteration}
 \EndIf   
 \State $W \gets M \cdot_J V$;\quad $M \gets M - \beta\, W\cdot_{R}V^\top$; \quad $U \gets \texttt{ColNorm}(W)$

\State \textbf{return} $\sqrt{|I|/|J|}\,U\cdot_RV^\top,\; M,\; V$
\EndFunction
\end{algorithmic} 

\vspace{-0.8em}
\rule{\textwidth}{0.4pt}
\vspace{-1.3em}

\begin{algorithmic}[1]
\Function{$\algname{CPU-Dion}^{0\mathrm{D}}$}{$\mbox{gradient }G[I,J],\,M[I,J],\,V[J,R];\, \future$}
\State $M \gets M + G$   

\If{$\future.\texttt{ready}()$} \algcomment{CPU job done}
\State $V \gets \future.\texttt{get}()$ \algcomment{refresh $V$ from CPU power iteration}
\State $\future \gets \texttt{SpawnCPU}(\algname{SymPowerIter1}(M,V))$ \algcomment{launch new job on CPU}
\EndIf 
 \State $W \gets M \cdot_J V$;\quad $M \gets M - \beta\, W\cdot_{R}V^\top$; \quad $U \gets \texttt{ColNorm}(W)$   
\State \textbf{return} $\sqrt{|I|/|J|}\,U\cdot_RV^\top,\; M,\; V,\; \future$
\EndFunction
\end{algorithmic}

\vspace{-0.8em}
\rule{\textwidth}{0.4pt}
\vspace{-1.3em}

\begin{algorithmic}[1]
\Function{$\algname{SymPowerIter1}$}{$M[I,J],\,V[J,R]$} \algcomment{single `symmetric' power iteration}
\State $V[J,R] \gets  M^\top \cdot_I M \cdot_J V$
\State $V[J,R] \gets \texttt{Orthonormalize}(V)$  
\State \textbf{return} $V$
\EndFunction
\end{algorithmic}
\end{algorithm}

\subsection{Dion Variants for Extreme DP Communication Constraints}
\label{sec:double-dion-main}

Under hybrid sharding scenarios where inter-node bandwidth is severely constrained,  Dion can be adapted to minimize replicated-DP communication. 
\textbf{Double Dion} (see \autoref{alg:double-dion}) is a two-stage variant: Stage~1 performs a DP-synchronized update at a very small rank $r_1 \ll r$, while Stage~2 completes the update locally at a larger rank $r_2$ without any DP communication. 
This design reduces the DP payload to $(m{+}n)r_1$ while still benefiting from a higher-rank local update. 
The two stages deliberately use distinct error-feedback rules, which we find essential for stability. 
A further variant overlaps the Stage~1 all-reduce with forward/backward passes by applying a one-step delayed sketch, yielding nearly free communication at the cost of a slight slowdown in convergence. 
Full details and empirical results are provided in \autoref{sec:double-dion}.

\subsection{Effective Ranks Filtering}

Not all directions in the momentum matrix $M$ contribute equally to 
learning. Empirically, Dion updates span only a moderate number of 
informative directions: the effective rank~\citep{roy2007effective}, 
$2^{\mathrm{erank}(M)}$, typically lies between 400--800 for hidden dimension 1024. Many columns of $W$ therefore capture low-energy directions that add cost but little benefit.
Effective rank also varies systematically—higher in middle layers, rising 
sharply early in training, and shrinking during learning rate decay—suggesting 
that the ``useful'' rank is both smaller and nonstationary. This motivates two 
adaptive strategies: (i) filtering small-norm columns of $W$, and (ii) setting 
the update rank to the exponential effective rank. In practice, filtering 
matches full-rank accuracy at lower cost, while effective-rank adaptation 
remains competitive with only a minor gap (\autoref{sec:update_rank}).

\section{Experimental Results}
\label{sec:exp}

We present a range of experimental results for Dion, comparing to baselines of AdamW~\citep{loshchilov2018decoupled} and Muon~\citep{jordan2024muon}.
All experiments are conducted using GPT-style decoder-only Transformer models at various scales, trained on the FineWeb dataset~\citep{penedo2024the} or the FineWeb-Edu dataset~\citep{lozhkov2024fineweb-edu}, following the setup of the \texttt{modded-nanoGPT} codebase~\citep{modded_nanogpt_2024}. 
For additional details on training configurations and hyperparameters, please refer to \autoref{sec:hyper}.

\subsection{Larger Models Tolerate Lower Ranks}
\label{sec:model}

\begin{figure}
    \centering
    \includegraphics[width=\linewidth]{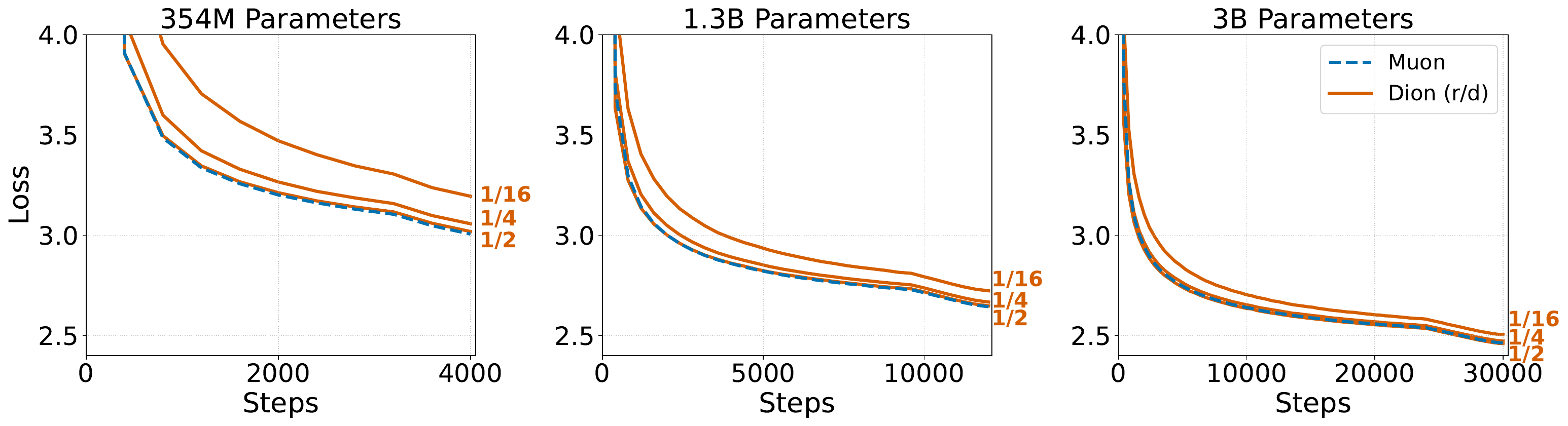}
    \caption{\footnotesize
   Validation loss for Dion at three low-rank settings ($r=d/2,\;d/4,\;d/16$) versus full-rank Muon.  
Larger models are increasingly robust to low-rank update.
}
    \label{fig:scale}
\end{figure}
\autoref{fig:scale} reports validation loss curves across three model scales. We compare Dion with rank fractions $r/d \in \{1/2,\,1/4,\,1/16\}$, where $r$ is the update rank and $d$ the hidden dimension. \textbf{Larger models are more tolerant of lower ranks:} at 3B parameters, Dion with $r=d/2$ or $d/4$ performs on par with full-rank Muon, and the gap between $r=d/16$ and $r=d/2$ narrows substantially. Full experimental details are provided in \autoref{details:model}.

\subsection{Preserved Benefits in Large-Batch Training}
\label{sec:largebatch}

Muon is well known for its suitability in large-batch training, achieving favorable compute–time tradeoffs by retaining data efficiency even at very large batch sizes~\citep{shah2025practical}.  
To verify that Dion inherits this property, we conducted small-scale experiments with a 160M-parameter model (\autoref{fig:cbs}), following the critical batch size protocol of~\citet{zhang2025how}.  
The results show that Dion preserves Muon’s robustness to large batches, while remaining competitive with AdamW even at reduced ranks ($r/d = 1/8$).  
Full experimental details are provided in \autoref{sec:cbs}.

\subsection{Hyperparameter Transfer}
\label{sec:transfer}

\begin{figure}
    \centering
    \includegraphics[width=0.85\linewidth]{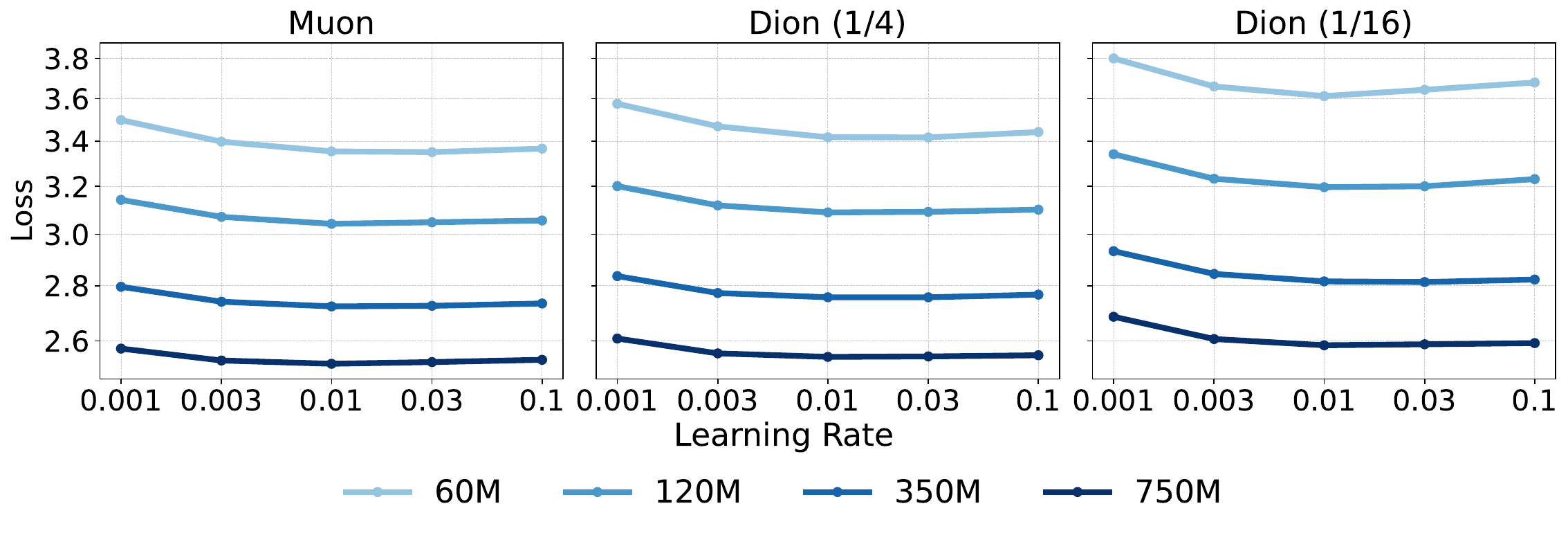}
    \vspace{-10pt}
    \caption{ \footnotesize Learning rate versus loss for four model scales. Dion maintains hyperparameter transferability trends of Muon across both model size and different rank fractions.}
    \label{fig:transfer}
\end{figure}

Previous work~\citep{bernstein2025deriving, bernstein2024modular, pethick2025training} has shown that Muon can exhibit learning rate transfer across model size via shape-dependent scale factors (\autoref{sec:scalar}). To empirically verify that Dion inherits this property, we sweep learning rates for Muon and Dion with rank fractions $1/4$ and $1/16$ across four model sizes, each trained for a Chinchilla-optimal number of tokens \citep{hoffmann2022training}. As shown in \autoref{fig:transfer}, optimal learning rates are approximately identical for all model sizes. Detailed hyperparameters are given in \autoref{details:transfer}.
\subsection{Faster Dion Variants}
\label{sec:fast-exp}

\begin{figure}[h]
    \centering
    \includegraphics[width=0.9\linewidth]{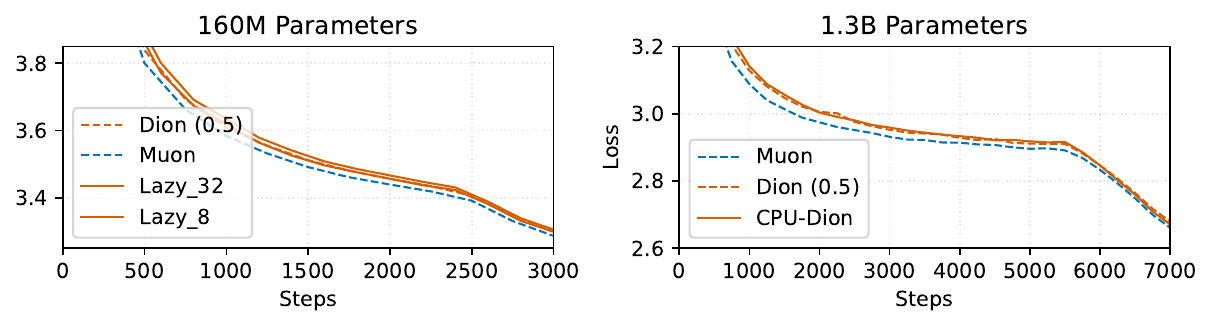}
    \vspace{-10pt}
    \caption{\footnotesize 
    Convergence of faster Dion variants. Both \algname{Lazy-Dion} and \algname{CPU-Dion} show only minor degradation relative to Muon (Dion at rank fraction $r/d=0.5$ included for comparison).}
    \label{fig:lazy}
\end{figure}

We evaluate the faster variants introduced in \autoref{sec:lazy}.  
\autoref{fig:lazy} reports results for \algname{Lazy-Dion} and \algname{CPU-Dion}, the latter simulated by introducing a one-step delay in updating $V$, corresponding to overlap between power iteration and a forward/backward pass.  
Both approaches show only minor convergence degradation compared to standard Dion.  
Notably, \algname{Lazy-Dion} with update frequency $k=32$ remains competitive with Dion at rank fraction $r/d = 0.5$, suggesting significant potential for further speedups at scale.

\subsection{Additional Experiments}

We present several supplementary studies.  
First, preliminary fine-tuning experiments with Dion are reported in \autoref{sec:fine-tuning}.  
Second, \autoref{sec:rank-fraction} compares different rank fractions and benchmarks Dion against the compressed-update optimizer DeMo~\citep{peng2024decoupled}.  
Third, ablation studies on key algorithmic components (power iteration and error-feedback) are provided in \autoref{sec:ablate}.  
Fourth, a preliminary speedrun experiment on the 350M model of~\citet{modded_nanogpt_2024} is presented in \autoref{sec:speed}.  
Finally, empirical studies on the update rank are given in \autoref{sec:update_rank}.

\section{Related Work and Conclusion}
\label{sec:conclu}

Prior work on preconditioned optimizer update rules includes Shampoo \citep{gupta2018shampoo} and SOAP \citep{vyas2025soap}. These second-order optimization algorithms have substantial compute and memory overheads, motivating the development of distributed \citep{anil2020scalable, shi2023distributed} and quantized \citep{wang20244bit} implementations of Shampoo. COSMOS~\citep{liu2025cosmos} has shown promising results from combining SOAP- and Muon-style updates across different eigenspaces.

Prior approaches to reducing communication overhead include gradient sparsification~\citep{wang2018atomo,wang2023cocktailsgd} and federated averaging~\citep{mcmahan2017communication, douillard2023diloco}. We consider these techniques as complementary to Dion. Dion aims to be an efficient optimizer in itself, and it can be combined with these techniques to further lower communication.

Dion also connects to recent studies on low-rank updates~\citep{cosson2023low,jadbabaie2023adaptive} and memory-efficient optimizers like GaLore~\citep{zhao2024galore}. In contrast to observations from~\cite{song2024does}, the results here suggest that low-rank training can succeed when paired with an effective error feedback rule.
 
Lastly, we highlight a few directions for future work:

\vspace*{-5pt}
\begin{itemize}[noitemsep]
    \item \textbf{Quantization:} Dion’s optimizer states may be quantizable to lower precision formats to reduce memory use. The column-normalized $Q$ matrix may be particularly quantization-friendly. In addition, expensive steps like QR decomposition may also be faster in reduced-precision arithmetic.
    \item \textbf{Error feedback:} Refining the error feedback rule may improve convergence at lower ranks. A variant explored in \autoref{sec:double-dion} shows promising results in this regime.
    \item \textbf{Beyond LLMs:} While experiments here focus on GPT-style models, any architecture with matrix-shaped parameters and dense activation vectors may benefit from orthonormalized updates.
\end{itemize}

\section*{Acknowledgments}

We are grateful to Laker Newhouse and Jeremy Bernstein for their many constructive suggestions on the manuscript.
We also thank Shital Shah for his helpful feedback and assistance.

\bibliographystyle{plainnat}
\bibliography{refs}

\newpage
\appendix

\renewcommand{\appendixpagename}{\centering \LARGE Appendix}
\appendixpage

\startcontents[section]
\printcontents[section]{l}{1}{\setcounter{tocdepth}{2}}
\newpage

\section{Accelerating Power Iteration with Cholesky QR}

\label{sec:cholesky}

\texttt{Orthonormalize} is a key step in the amortized power iteration of Dion, ensuring accurate recovery of the top-$r$ singular vectors. Our default choice is the standard QR decomposition (\texttt{torch.linalg.qr}) or the RCQR algorithm (see \autoref{alg:dist-orth}), both of which internally rely on QR. While QR via Householder reflections is numerically stable, it is also computationally expensive.

Cholesky QR (CQR) provides a much faster alternative. \autoref{fig:exectime_ratio} reports the speedup of CQR relative to QR and RCQR across different matrix sizes and rank fractions. We find that CQR remains numerically stable for matrices with condition number below $5 \times 10^3$. Beyond this threshold, however, CQR may suffer from instabilities and precision loss. In practice, our initial experiments showed occasional training instabilities and loss spikes when using CQR directly.

\begin{table}[H]
\centering
 
\begin{minipage}{0.45\textwidth}
  \centering
  \scriptsize
\begin{tabular}{c|cccccc}
\hline
\multirow{2}{*}{Dimension} & \multicolumn{6}{c}{Rank Fraction} \\
 & 1.0 & 0.5 & 0.25 & 0.125 & 0.0625 & 0.0312 \\
\hline
768 & 3.64 & 4.14 & 4.41 & 3.73 & 2.74 & 2.01 \\
1024 & 3.83 & 4.68 & 4.64 & 4.46 & 5.00 & 2.85 \\
2048 & 3.41 & 4.47 & 5.34 & 5.19 & 5.07 & 4.78 \\
4096 & 3.29 & 5.32 & 6.82 & 8.58 & 8.30 & 8.36 \\
8192 & 2.50 & 4.20 & 6.09 & 8.35 & 10.39 & 11.43 \\
16384 & 1.29 & 2.79 & 4.25 & 6.03 & 8.38 & 10.16 \\
\hline
\end{tabular}
\captionsetup{labelformat=empty}
\captionof{table}{$\frac{\text{QR}}{\text{CholeskyQR}}$ execution time ratio.}
\end{minipage} \quad \quad \quad
\begin{minipage}{0.45\textwidth}
\centering
  \scriptsize
 \begin{tabular}{cccccc}
\hline
    \multicolumn{6}{c}{Rank Fraction} \\
   1.0 & 0.5 & 0.25 & 0.125 & 0.0625 & 0.0312 \\
\hline
  4.64 & 4.39 & 3.76 & 5.84 & 4.79 & 2.46 \\
  4.13 & 4.53 & 4.11 & 3.91 & 6.28 & 3.13 \\
  5.14 & 5.57 & 4.56 & 4.19 & 3.73 & 6.41 \\
  5.25 & 4.73 & 4.96 & 4.41 & 4.13 & 3.83 \\
  4.44 & 4.40 & 4.14 & 3.18 & 3.96 & 3.93 \\
 3.65 & 3.55 & 3.09 & 3.54 & 3.51 & 3.44 \\
\hline
\end{tabular}
\captionsetup{labelformat=empty}
\captionof{table}{$\frac{\text{RCQR}}{\text{CholeskyQR}}$ execution time ratio.}
\end{minipage}
\caption{Speedup of CQR relative to QR (left) and RCQR (right). Each entry corresponds to a $(d, d \cdot \alpha)$ matrix with dimension $d$ and rank fraction $\alpha$.}

\label{fig:exectime_ratio}
\end{table}

\begin{figure}[h]
    \centering
    \includegraphics[width=\linewidth]{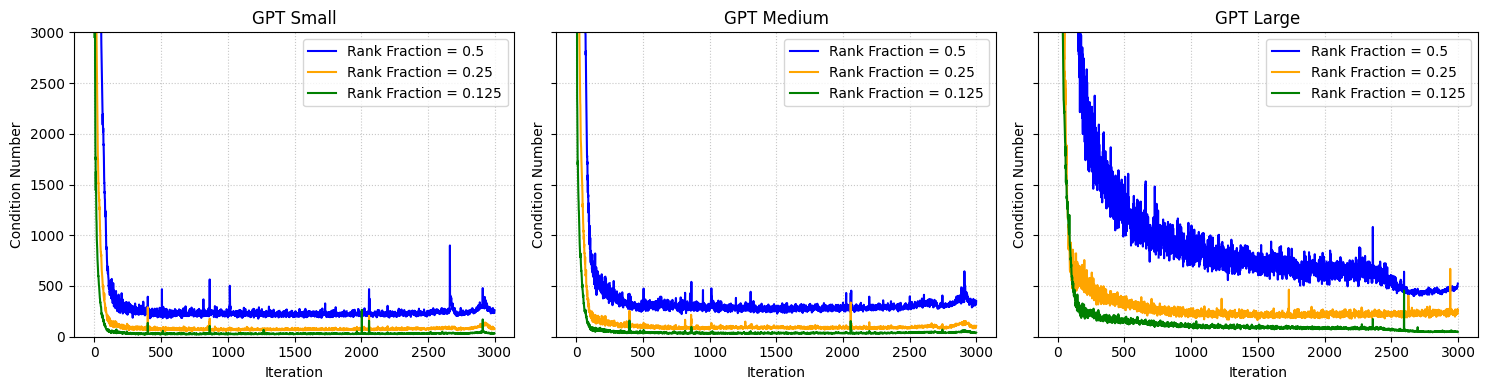}
    \caption{\footnotesize
    Maximum condition number of $P$ before orthogonalization, across training iterations.  
}
    \label{fig:condnumber_scale}
\end{figure}

\begin{figure}[h]
    \centering
    \includegraphics[width=\linewidth]{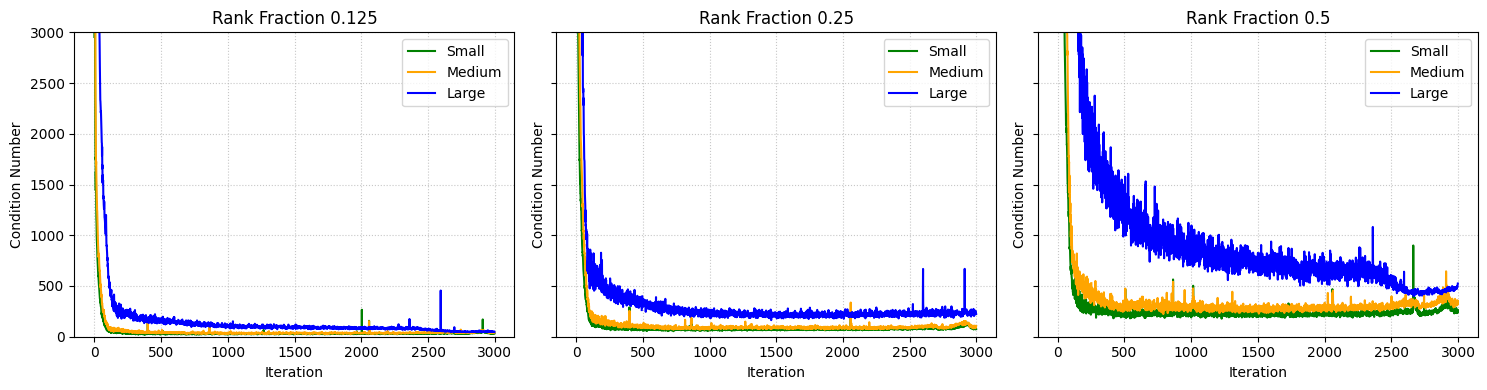}
    \caption{\footnotesize
    Maximum condition number of $P$ before orthogonalization, as a function of rank fraction.  
}
    \label{fig:condnumber_rf}
\end{figure}

\paragraph{Condition number dynamics.}  
To better understand when CQR can be applied safely, we examined the condition number of $P$ prior to orthonormalization. As shown in \autoref{fig:condnumber_scale}, the condition number decreases and stabilizes after a few hundred iterations, allowing us to switch from QR to CQR after a warm-up period. We observe:
\begin{itemize}
\item At fixed model size, the equilibrium condition number decreases as the rank fraction decreases (\autoref{fig:condnumber_scale}).
\item At fixed rank fraction, the equilibrium condition number increases with model size (\autoref{fig:condnumber_rf}). 
\end{itemize}
This suggests that for larger models, using smaller rank fractions improves CQR stability. Outlier layers (top $10\%$ of condition numbers) were excluded, since rare failures are already handled by our fallback mechanism.

\subsection{Why the condition number stabilizes}
The drop in condition number reflects Dion’s ability to learn the subspace of dominant singular vectors of the momentum matrix $M$. Although $M$ itself remains ill-conditioned during training, the projection $P \leftarrow M V$ becomes well-conditioned because $V$ rapidly aligns with the top right singular vectors of $M$, filtering out smaller singular directions. Lower rank fractions accentuate this effect, explaining the stronger stabilization seen at $\alpha=0.25$ or $\alpha=0.125$.

\subsection{Implementation details}
During training, we found that CQR may occasionally fail for certain matrices. To handle this, we utilize the functionality provided by \texttt{torch.cqr}, which reports failure cases. When a failure is detected, we fall back to the standard orthogonalization method. Since such failures occur very infrequently (less than 1\% of the time), the overhead of retrying with the fallback method is negligible.

\section{Additional Experiments}

\subsection{Preserved Benefits in Large-Batch Training}
\label{sec:cbs}

\begin{figure}[h]
    \centering
    \includegraphics[width=0.45\linewidth]{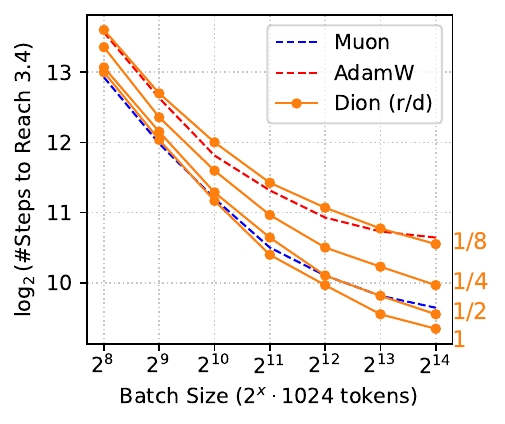} 
    \caption{\footnotesize 
     Comparison of Dion, Muon, and AdamW across batch sizes $2^{8}$–$2^{14}$ sequences per batch (each sequence has $1024$ tokens).  
    Notably, Dion preserves Muon’s ability to tolerate very large batch sizes, while remaining competitive with AdamW even at reduced ranks.}
    \label{fig:cbs}
\end{figure}

We adopt the critical batch size evaluation protocol of~\citet{zhang2025how}, with one modification: instead of applying an exponential moving average, we use a fixed learning rate for each optimizer and relax the target loss threshold to allow fairer comparison. All experiments use a 160M-parameter model trained on FineWeb until reaching a validation loss of $3.4$.

For AdamW, we sweep learning rates $\{0.001, 0.003, 0.01\}$, weight decay $\{0, 0.01\}$, and $\beta_2 \in \{0.95, 0.99\}$, using a trapezoidal learning rate schedule with $10\%$ warmup and $10\%$ cooldown, and fixing $\beta_1 = 0.9$.  
For Muon and Dion, we fix the momentum parameter $\mu = 0.95$ and sweep learning rates over $\{0.003, 0.01, 0.03\}$.

\autoref{fig:cbs} shows that Dion’s performance degrades gracefully as the rank $r$ decreases.  
Across all ranks tested, Dion maintains a critical batch size at least as large as Muon’s and consistently larger than AdamW’s.  
This suggests that Dion inherits Muon’s robustness to large batch training, while offering flexibility through adjustable update rank.

\subsection{Preliminary Experiments on Fine-Tuning}
\label{sec:fine-tuning}
 
\begin{figure}[H]
    \centering
    \begin{subfigure}[b]{0.45\textwidth}
        \centering
        \includegraphics[width=\linewidth]{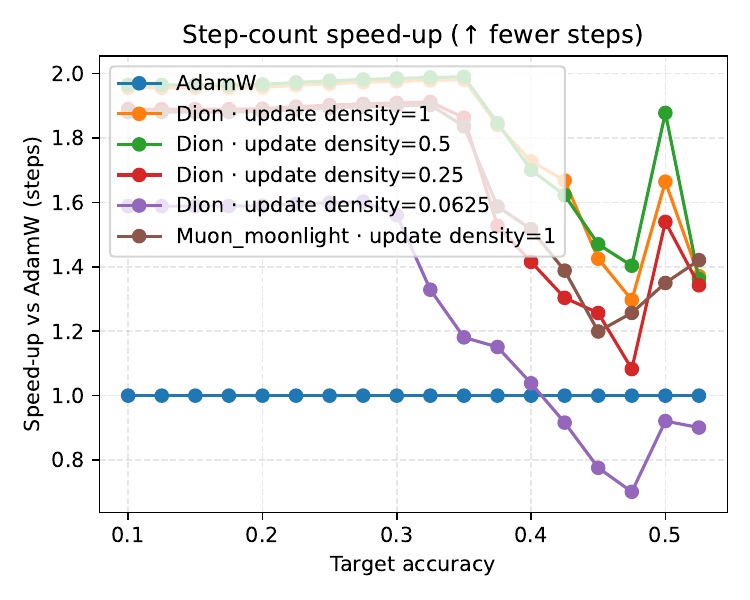}
        \caption{}
        \label{fig:a}
    \end{subfigure}
    \hfill
    \begin{subfigure}[b]{0.45\textwidth}
        \centering
        \includegraphics[width=\linewidth]{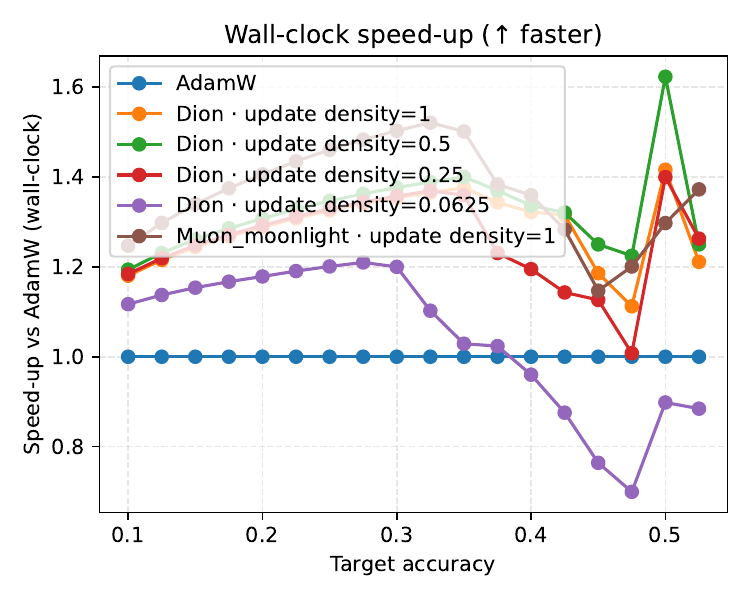}
        \caption{}
        \label{fig:b}
    \end{subfigure}
    
    \begin{subfigure}[b]{0.45\textwidth}
        \centering
        \includegraphics[width=\linewidth]{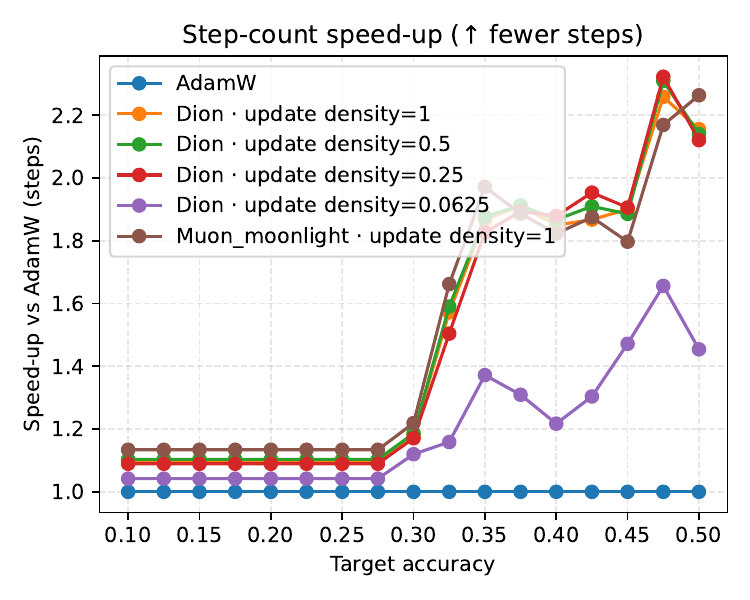}
        \caption{}
        \label{fig:c}
    \end{subfigure}
    \hfill
    \begin{subfigure}[b]{0.45\textwidth}
        \centering
        \includegraphics[width=\linewidth]{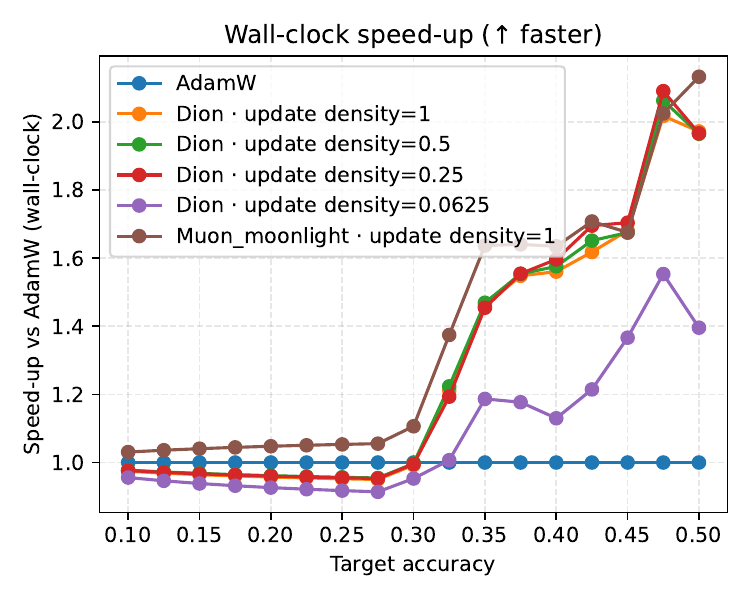}
        \caption{}
        \label{fig:d}
    \end{subfigure}
    
    \caption{\textbf{Optimization efficiency in model adaptation.}  
    Dion converges faster than Adam in both full fine-tuning (\emph{top}: a, b) and LoRA ($r=256$) (\emph{bottom}: c, d), while matching Muon’s performance. The embedding dimension is 2048.}
    \label{fig:finetuning}
\end{figure}

We now test whether Dion’s benefits extend to fine-tuning.  
Specifically, we adapt a pre-trained decoder-only model, Gemma 2B~\citep{team2024gemma}, on curated subsets of OpenMathInstruct-2~\citep{toshniwal2025openmathinstruct}, a high-quality dataset derived from GSM8K and MATH, and evaluate on GSM8K~\citep{cobbe2021training}.  
We compare Dion, Muon, and Adam under two adaptation strategies: (i) full fine-tuning and (ii) Low-Rank Adaptation (LoRA)~\citep{hu2022lora}. The model’s hidden dimension is $2048$.

As shown in \autoref{fig:finetuning}, Dion demonstrates some efficiency gains over Adam. In the full fine-tuning setting, Dion achieves a 1.5× reduction in wall-clock time and a 1.8× speed-up in step count. For LoRA-based fine-tuning, the improvements are even more pronounced, with Dion offering a 2× reduction in wall-clock time and a 2.2× speed-up in step count compared to Adam. Notably, both Muon and Dion exhibit similar effectiveness during fine-tuning.

Additionally, Dion enables effective training of lower-rank LoRA adapters ($r=128$ and higher) while achieving performance comparable to full fine-tuning. In contrast, Adam requires higher-rank LoRA configurations ($r=256$ and higher) to reach similar levels of accuracy. This result is particularly significant because low-rank adaptations like LoRA are designed to reduce the number of trainable parameters during fine-tuning, thereby lowering memory usage and computational cost. However, training effectively with very low-rank parameterizations is often challenging due to optimization difficulties, such as limited representational capacity and slower convergence. Dion’s ability to achieve comparable performance with lower-rank LoRA modules suggests that it improves the optimization landscape or gradient dynamics, allowing the model to learn efficient parameter updates even under tighter constraints.

\subsection{Performance Across Rank Fractions}
\label{sec:rank-fraction}

\begin{figure}[H]
    \centering
\includegraphics[width=0.4\linewidth]{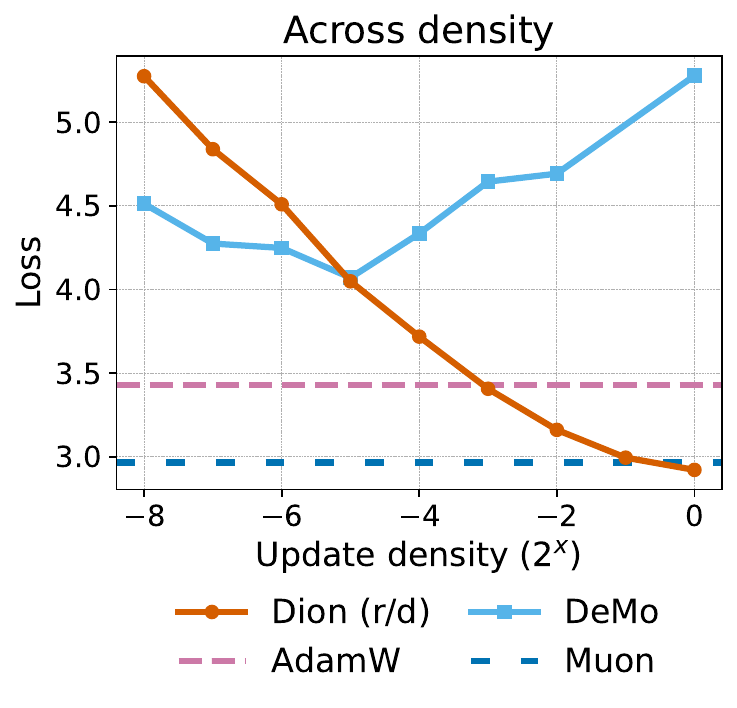}
    \caption{\footnotesize
    Dion compared with DeMo, Muon, and AdamW as the rank fraction $r/d$ increases. Dion improves steadily with higher update density, matching DeMo at $r/d = 1/32$ and AdamW at $r/d = 1/8$.}
    \label{fig:density-batch}
    \vspace{-10pt}
\end{figure}

Dion reduces communication by lowering the rank fraction $r/d$. For comparison, we include DeMo \citep{peng2024decoupled}, which compresses gradients via the Discrete Cosine Transform (DCT) with two hyperparameters: top-$k$ frequency selection and chunk size $s \times s$. To make results comparable, we define the \emph{update density} as the fraction of retained degrees of freedom: $\delta = r/d$ for Dion and $\delta = k/s^2$ for DeMo. AdamW and Muon perform uncompressed updates, corresponding to $\delta = 1$.

In \autoref{fig:density-batch}, Dion remains competitive across moderate compression. It consistently surpasses DeMo once $\delta > 1/32$, exceeds AdamW from $\delta \geq 1/8$, and slightly outperforms Muon at full rank. All experiments use 160M-parameter models trained on FineWeb-Edu with sequence length $1024$ and batch size $4$M tokens.  

Dion achieves its strongest performance at full rank, with accuracy declining smoothly as $r/d$ decreases. DeMo shows a less stable profile: it peaks at $\delta = 1/32$ but degrades for both higher and lower densities. At sufficiently low $\delta$, DeMo can outperform Dion. We revisit this regime in \autoref{sec:double-dion} with a Dion variant designed for low-rank settings.

\textbf{Training setup.}  
All runs use a constant learning rate schedule with a 10\% linear cooldown. AdamW and DeMo additionally use a 10\% warmup. The hyperparameter configurations are:  
\begin{itemize}
    \item \textbf{AdamW:} $(\beta_1,\beta_2)=(0.9,0.95)$; learning rates $\{0.001, 0.003, 0.01\}$; weight decay $\{0, 0.01\}$.  
    \item \textbf{Muon and Dion:} momentum $\mu=0.95$; learning rates $\{0.003, 0.01, 0.03\}$; inner Adam scalar optimizer with learning rate $0.002$ and $(\beta_1,\beta_2)=(0.9,0.95)$.  
    \item \textbf{DeMo:} learning rates $\{0.0003, 0.001, 0.003\}$; compression decay $0.999$.  
\end{itemize}

Both Dion and DeMo require roughly $2\delta$ of the full gradient size in data-parallel synchronization. Dion exchanges two dense matrices of size $m \times r$ and $n \times r$, while DeMo communicates sparse updates with $k$ indices and $k$ values per chunk ($k/s^2$ density). The exact cost depends on the model shape for Dion and the sparse representation for DeMo.

\subsection{Ablation Studies}
\label{sec:ablate}

We conduct ablation studies to evaluate two core components of Dion: the error feedback mechanism and the use of a single power iteration for low-rank approximation. For these ablation studies, we train 160M parameter models and set the batch size to be $2048 \cdot 1024 \approx 2.1$~M tokens.

\textbf{Single power iteration vs. full SVD.}
We evaluate the effectiveness of Dion’s single-step power iteration for computing a rank-$r$ approximation. In this experiment, we compare Dion to an alternative that computes the singular value decomposition (SVD) at every step and truncates to the top-$r$ singular values---an ideal but computationally expensive alternative.

\begin{figure}[H] \centering \begin{subfigure}{0.45\linewidth} \centering \includegraphics[width=\linewidth]{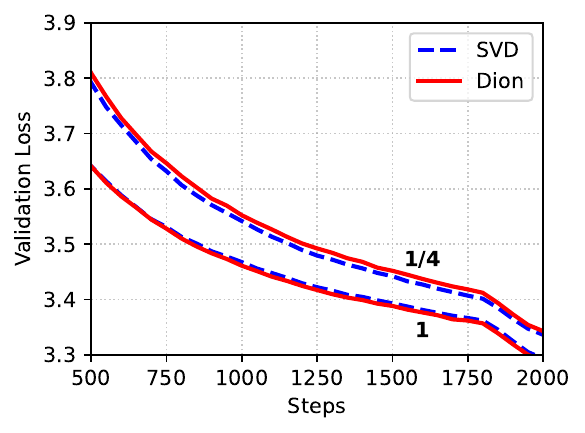} \caption{Ablating orthonormalization} \label{fig:svd} \end{subfigure} 
\begin{subfigure}{0.45\linewidth} \centering \includegraphics[width=\linewidth]{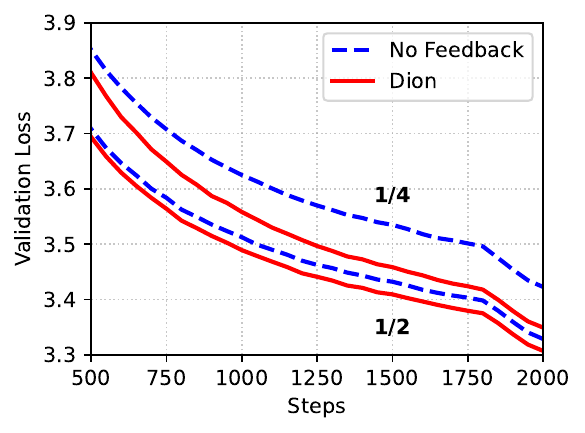} \caption{Ablating error feedback} \label{fig:error} \end{subfigure} 

\caption{\footnotesize \textbf{Testing algorithmic components.} We evaluate two key design choices in Dion: the error feedback mechanism and the use of a single step of power iteration. Error feedback proves crucial at lower ranks. A single power iteration performs on par with SVD, offering a more efficient alternative without sacrificing performance.}
\end{figure}

The results, shown in \autoref{fig:svd}, reveal negligible differences in convergence behavior between the two approaches. This suggests that using power iteration---initialized from the previous iteration’s right orthonormal basis \citep{vogels2019powersgd}---can give a sufficiently accurate approximation at a fraction of the computational cost.

\textbf{Importance of error feedback.}
To evaluate the necessity of the error feedback mechanism, we compare Dion against a simplified variant that omits error feedback. In this baseline, the rank-$r$ approximation is applied directly to the momentum buffer: \begin{align*} M \gets \mu M + G, \quad \text{for } \mu \in (0, 1), \end{align*}
whereas Dion  updates the momentum using error feedback,

As shown in \autoref{fig:error}, the version without error feedback suffers from steep performance degradation as the rank decreases from $r = d/2$ to $r = d/4$, while Dion maintains stable performance. This highlights the importance of incorporating error feedback to preserve optimization quality when using low-rank approximation.

\subsection{Results for the 350M Speedrun Configuration}
\label{sec:speed}

We also present the result using the 350M speedrun configuration in \url{https://github.com/KellerJordan/modded-nanogpt} with BF16 training and several changes to the model architecture. Following the 04/22/25 record, we add the Dion optimizer on top, perform basic hyperparameter tuning and end up with a slightly larger learning rate 0.03 and keep all the other hyperparameters.

\begin{table}[h]
\centering
\begin{tabular}{c|lll}
\toprule
& Muon                      & Dion (full rank)                       & Dion (75\% rank)                    \\
\hline
Validation loss & 2.920 &2.921&\textbf{2.919}\\

\bottomrule
\end{tabular}
\vspace{1mm}
\caption{350M speedrun results, following the configuration in \url{https://github.com/KellerJordan/modded-nanogpt}.}
\label{tab:speedrun}
\end{table}

We present our result in \autoref{tab:speedrun}.  We observe modestly better update quality from the 75\% rank, aligning with our finding in \autoref{sec:update_rank} that full-rank update is not always the best. However, we do observe 14.5\% relative worse wall-clock time from Dion due to QR being a more expensive operation than Newton-Schulz at the 350M parameter scale.

\subsection{Dion Variants for Extreme DP Communication Constraints}
\label{sec:double-dion}

\begin{algorithm}[t]
\caption{Double Dion (2D-sharded) on $[L,I_{\texttt{Y}},J_{\texttt{X}}]$ with ranks $|R_1|{=}r_1$ and $|R_2|{=}r_2$}
\label{alg:double-dion}
\begin{algorithmic}[1]
 
\Statex \textbf{Optimizer states:} 
\begin{itemize}
    \item \(\mbox{Stage-1 momentum }M_1[L,I_{\texttt{Y}},J_{\texttt{X}}]\) \emph{(decoupled across \texttt{Z})}, 
    \item 
\(\mbox{Stage-2 momentum }M_2[L,I_{\texttt{Y}},J_{\texttt{X}}]\) \emph{(synchronous across \texttt{Z})},
\item  \(\mbox{right factors }V_1[L,J_{\texttt{X}},R_1],\,V_2[L,J_{\texttt{X}},R_2]\).
\end{itemize}

\Statex \textbf{Hyperparameters:} error-feedback coefficients \(\beta_1,\beta_2 \in (0,1]\).

\Function{$\algname{Double\mbox{-}Dion}^{2\mathrm{D}}$}{$G,\,M_1,\,M_2,\,V_1,\,V_2$}

\State \textbf{Accumulate:}\quad \(M_1 \gets M_1 + G\)

\vspace{2pt}
\State \textbf{Stage 1 (small DP payload, rank $r_1$).}
\State \((U_1, W_1) \gets \algname{Distributed\mbox{-}PowerIter1}(M_1, V_1;\,\mathrm{DPsync}=\texttt{on})\)
\State \(M_1 \gets M_1 - \beta_1\, U_1\cdot_{R_1} W_1^\top\)
\State \(V_1 \gets \algname{ColNorm}(W_1)\)

\vspace{2pt}
\State \textbf{Stage 2 (no DP communication, rank $r_2$).}
\State \( M_2 \gets M_2 + U_1\cdot_{R_1} W_1^\top\) \algcomment{use DP-synchronized sketch}
\State \((U_2, W_2) \gets \algname{Distributed\mbox{-}PowerIter1}( M_2, V_2;\,\mathrm{DPsync}=\texttt{off})\)
\State \(M_2 \gets   M_2 - \beta_2\, U_2\cdot_{R_2} W_2^\top\)
\State \(V_2 \gets \algname{ColNorm}(W_2)\)

\vspace{2pt}
\State \textbf{Form update and return.}
\State \(O[L,I_{\texttt{Y}},J_{\texttt{X}}] \gets U_2\cdot_{R_2} V_2^\top\)
\State \textbf{return} \(\sqrt{|I|/|J|}\;O,\; M_1,\; M_2,\; V_1,\; V_2\)

\EndFunction
\end{algorithmic} 
\end{algorithm}

For hybrid sharding in very large-scale, geographically distributed training, it is often desirable to minimize replicated-DP communication while tolerating more relaxed FS and TP bandwidth. 
\textbf{Double Dion} is a two-stage variant: Stage~1 performs a DP-synchronized update at a very small rank $r_1 \ll r$, while Stage~2 completes the update locally at a larger rank $r_2$ without any DP communication. 
Stage~1 therefore communicates only $(m{+}n)r_1$ elements across \texttt{Z}, while Stage~2 operates exclusively along the model-parallel axes \texttt{X},\texttt{Y}. 
This makes Double Dion especially effective under extreme bandwidth constraints where full-rank DP synchronization is impractical.

In Stage~1, the per-replica momentum \(M_1\) is updated with the local gradient \(G\), and \(\algname{Distributed\mbox{-}PowerIter1}(M_1,V_1;\,\mathrm{DPsync}=\texttt{on})\) produces DP-synchronized factors \((U_1,W_1)\). Stage~1 then applies error feedback to \(M_1\) and refreshes \(V_1\).  

In Stage~2, the globally synchronous momentum \(M_2\) is updated with the synchronized sketch \(U_1 W_1^\top\), after which \((U_2,W_2)\) is obtained from \(\algname{Distributed\mbox{-}PowerIter1}(M_2,V_2;\,\mathrm{DPsync}=\texttt{off})\). Since all Stage~2 states are already synchronous across \texttt{Z}, no additional DP all-reduce is required. Distinct error-feedback coefficients are used: we find \((\beta_1,\beta_2)=(0.001,0.05)\) to be effective.  

\begin{figure}[h]
\centering
\includegraphics[width=\linewidth]{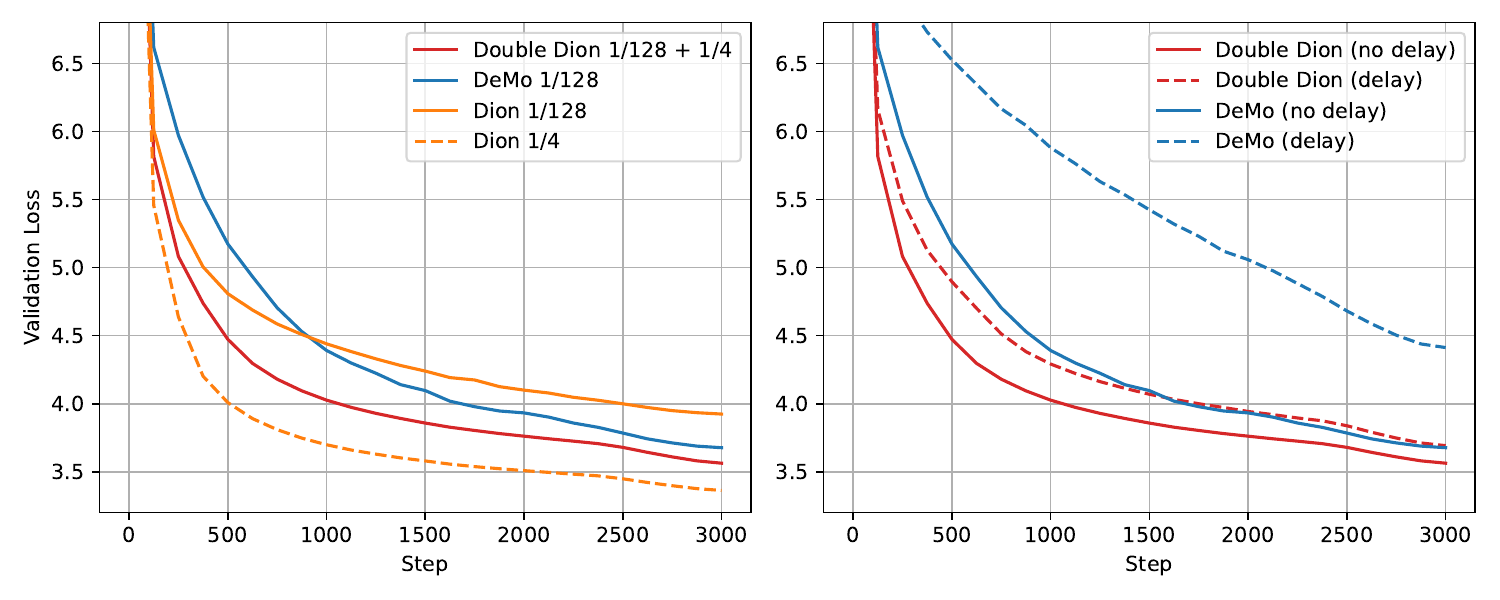}
\caption{\footnotesize \textbf{Left:} Comparison of Double Dion, Dion, and DeMo with equivalent DP communication at update density $1/128$. Dion with rank fraction $1/4$ is also shown for reference. Double Dion substantially improves upon standard Dion and outperforms DeMo. \textbf{Right:} Effect of introducing a one-step delay. Double Dion with delay matches DeMo without delay, while DeMo’s performance collapses with delay.}
\label{fig:double-dion}
\end{figure}

In \autoref{fig:double-dion} (left), we evaluate \autoref{alg:double-dion} with $r_1/d=1/128$ and $r_2/d=1/4$, comparing against DeMo with update density $1/128$ and standard Dion (\autoref{alg:2d-dion}) with update rank fractions $1/128$ and $1/4$. Note the extreme sparsity of $r_1/d=1/128$ for a 160M model: with $d_{\text{model}}=768$, this corresponds to only $r_1=6$.  

All three methods with $1/128$ update density incur the same DP communication volume, yet Double Dion achieves the lowest validation loss. We attribute this improvement to using two distinct error-feedback rules: attempts to apply a single rule to both stages yielded inferior results. The tradeoff is roughly doubled optimizer compute and memory cost. We view these results as preliminary and defer further refinement to future work.

\textbf{One-step delay (overlap).}  
Stage~2 can instead use the previous step’s $(U_1,W_1)$ to update \(M_2\), i.e.\ \(M_2 \gets M_2 + U_{1,t-1} W_{1,t-1}^\top\). This allows Stage~1 and Stage~2 to run in parallel and overlaps the Stage~1 DP all-reduce with the forward/backward pass. As shown in \autoref{fig:double-dion} (right), this delayed variant converges more slowly, but still outperforms Dion at $1/128$ and roughly matches DeMo without delay. In contrast, applying a one-step delay to DeMo’s compressed states severely degraded performance.

\textbf{Experimental setup.}  
We train 160M-parameter models on FineWeb with batch size 1024 and sequence length 1024 for 3000 steps, decaying the LR linearly to zero over the final 20\%. DeMo uses LR $0.001$. Dion and Double Dion both use LR $0.01$ with the scaling factors in \autoref{tab:scaling-factors}. Non-matrix parameters are optimized with Lion \((\beta_1,\beta_2)=(0.95,0.98)\), and one-step delay is applied only to Stage~1. Weight decay is $0.01$ for matrices and $0$ otherwise.

\subsection{Empricial Studies on the Update Rank}
\label{sec:update_rank}

\textbf{Effective rank.}
As proposed in~\citep{roy2007effective}, the concept of effective rank can be used to quantify the intrinsic dimensionality of matrix representations by measuring the entropy of their singular value distributions. The effective rank of the momentum matrix is:
\begin{equation}
    \text{erank}(M) = \exp\{H(p_1, p_2, ...,p_k)\},
\end{equation}
where $k = \min(m, n)$, $p_i = \frac{\sigma_k}{\Sigma_{i=1}^k|\sigma_k|}$ and $\sigma_i$ is the i-th singular value.  

As the effective rank is the entropy of the relative singular values, $2^{\text{erank}(M)}$ indicates the perplexity of the relative singular value distribution, i.e., how many directions the optimizer is effectively exploring. In practice, since the exact SVD is computationally expensive, we use an alternative which is readily available in Dion: the column norm of the $W$ matrix in \autoref{alg:un-dion}. Since $U$ is approximately orthonormal, the column norms of $W$ serve as an approximation of the singular values\footnote{In fact, we observe that the $L_2$ norm of the error of the estimated effective rank is consistently smaller than $0.1$ when comparing to the actual values calculated from SVD.}.

\begin{figure}[t]
    \centering
    
    \begin{subfigure}[b]{0.45\textwidth}
        \centering
        \includegraphics[width=\textwidth]{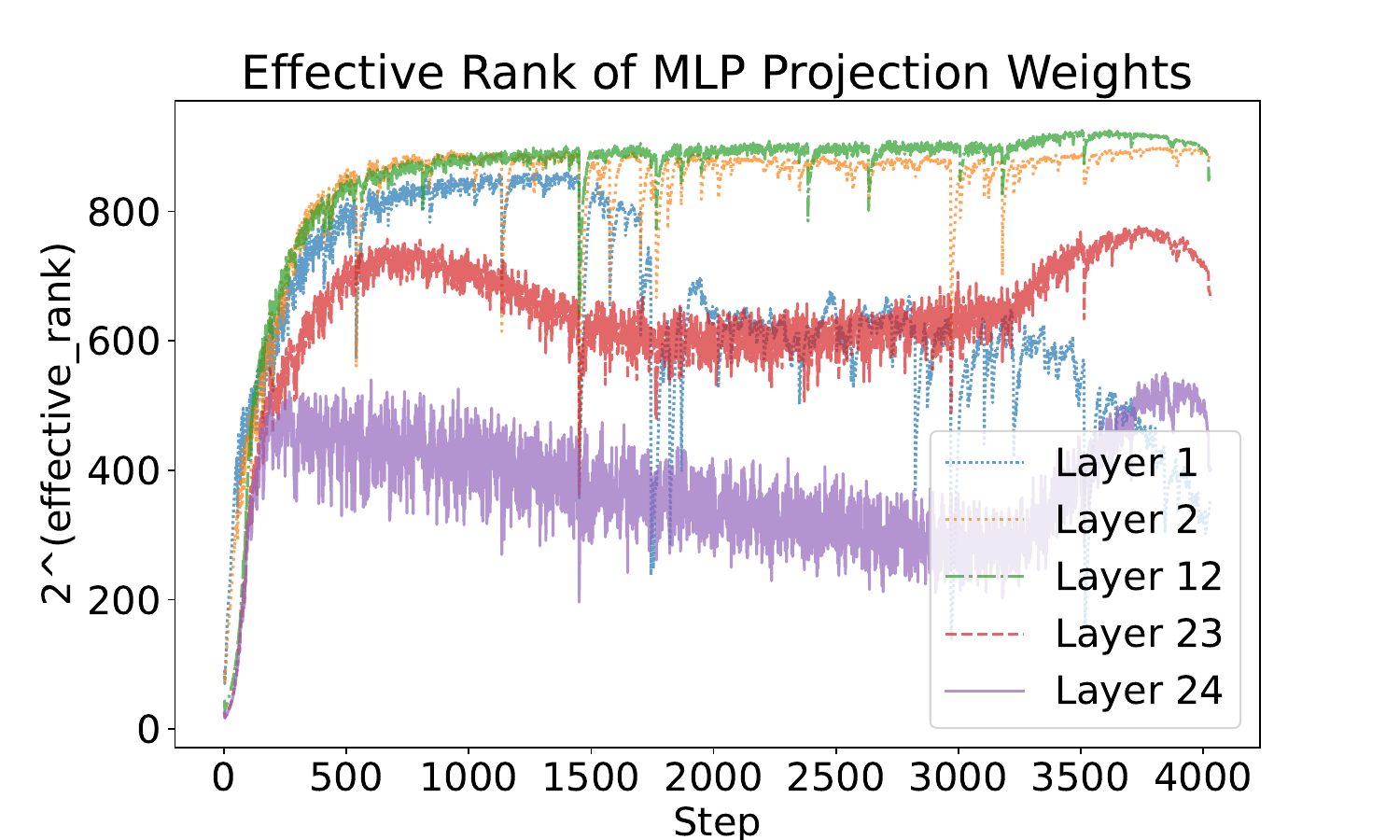}
        \caption{}
        \label{mlp_proj}
    \end{subfigure}
    \hfill
    \begin{subfigure}[b]{0.45\textwidth}
        \centering
        \includegraphics[width=\textwidth]{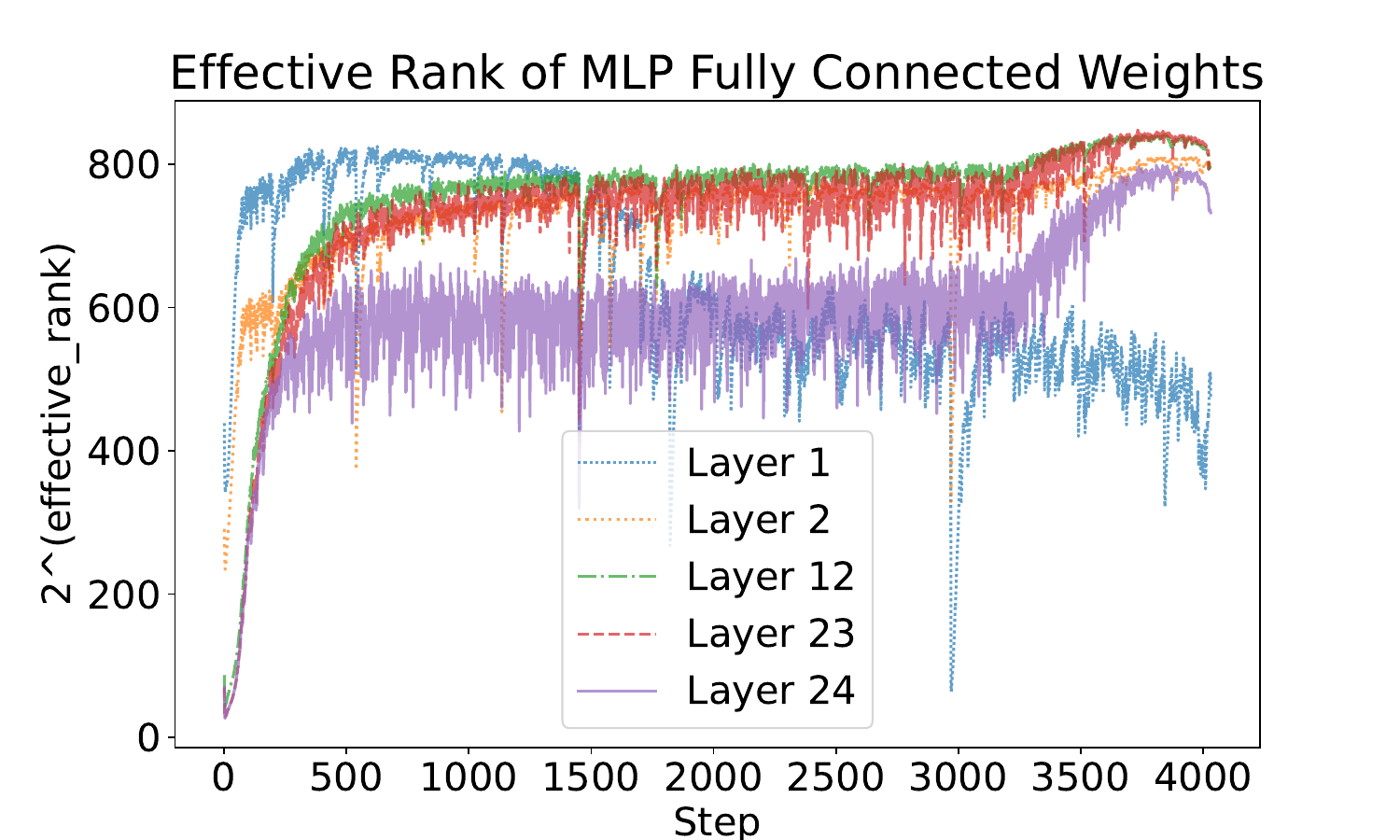}
        \caption{}
        \label{mlp_fc}
    \end{subfigure}

    \begin{subfigure}[b]{0.45\textwidth}
        \centering
        \includegraphics[width=\textwidth]{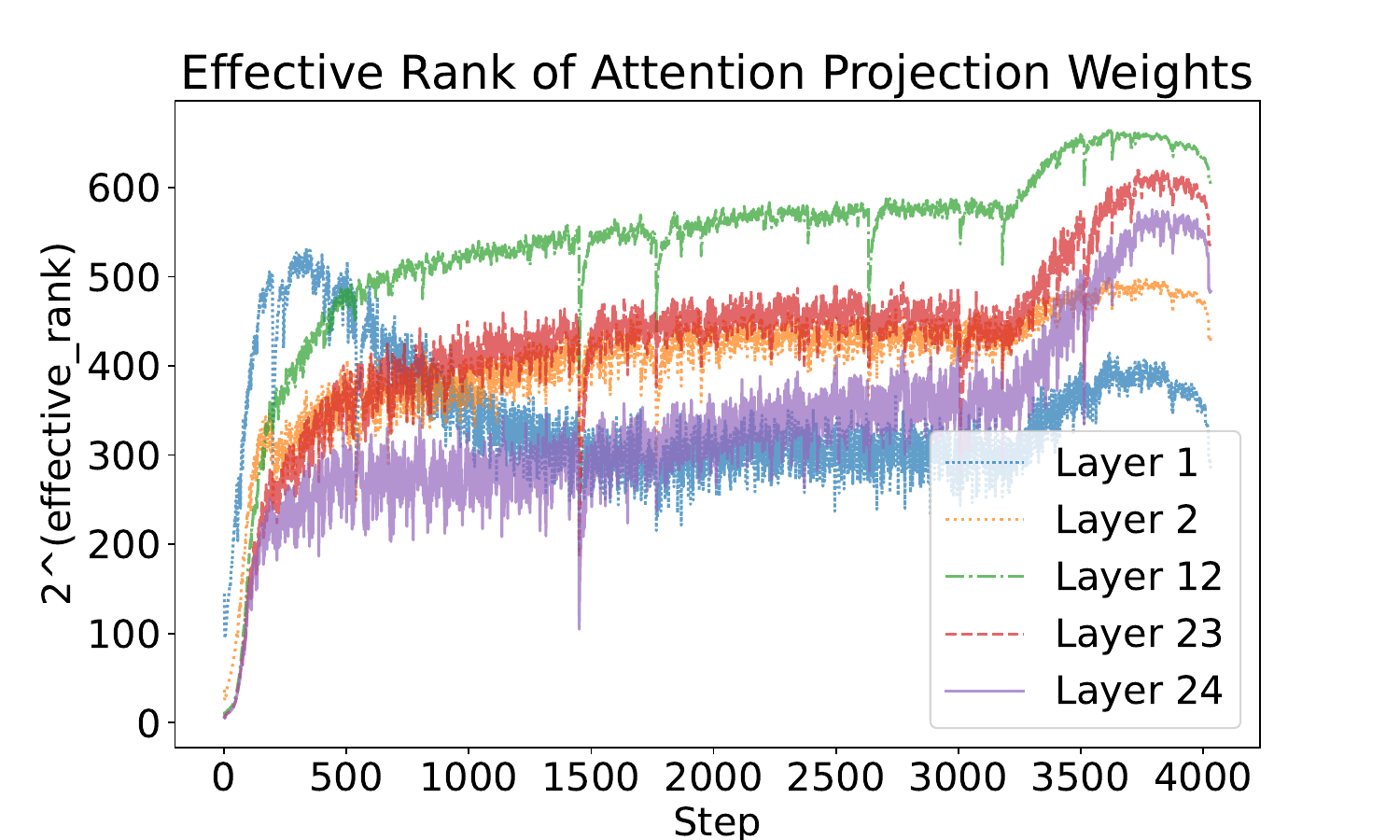}
        \caption{}
        \label{attn_proj}
    \end{subfigure}
    \hfill
    \begin{subfigure}[b]{0.45\textwidth}
        \centering
        \includegraphics[width=\textwidth]{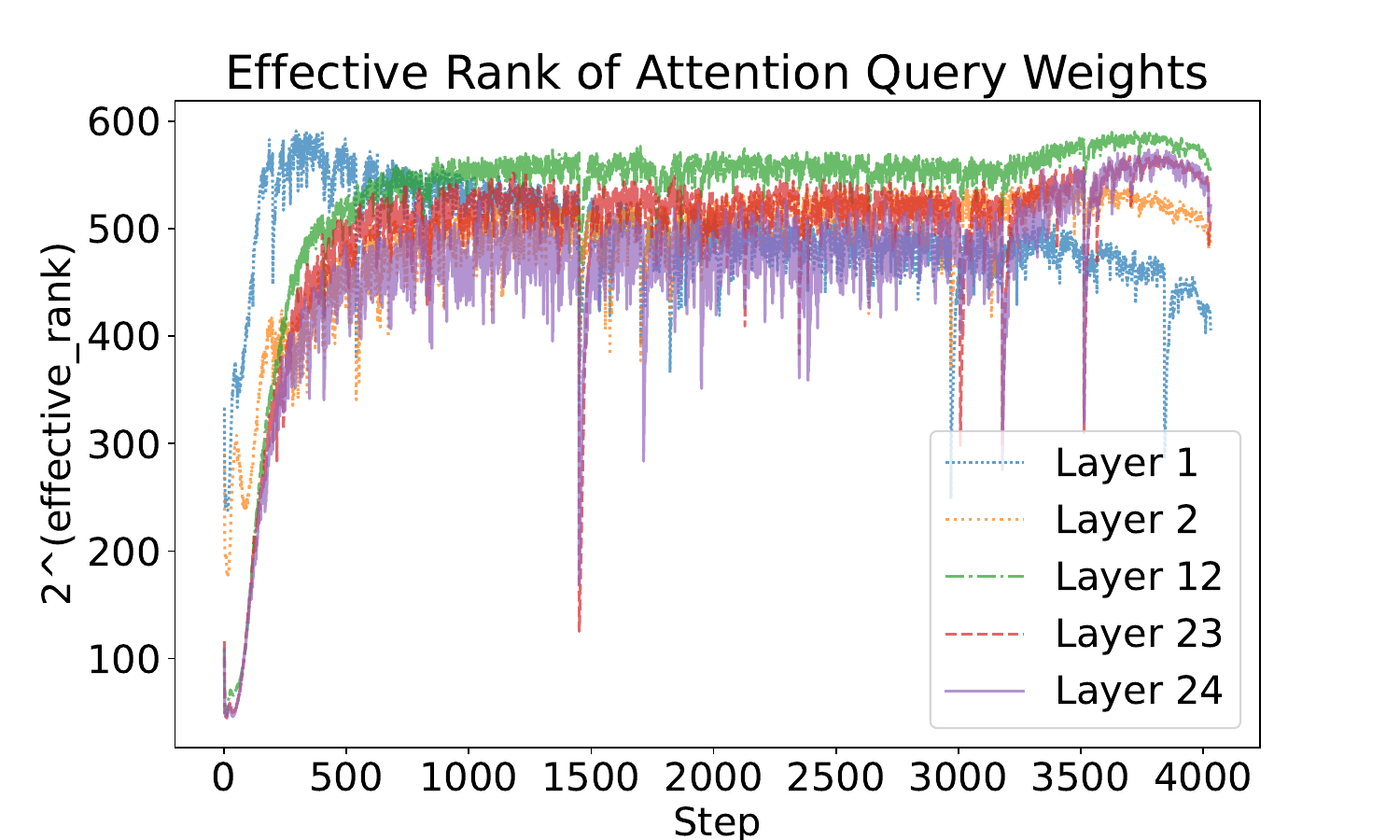}
        \caption{}
        \label{attn_q}
    \end{subfigure}

    \begin{subfigure}[b]{0.45\textwidth}
        \centering
        \includegraphics[width=\textwidth]{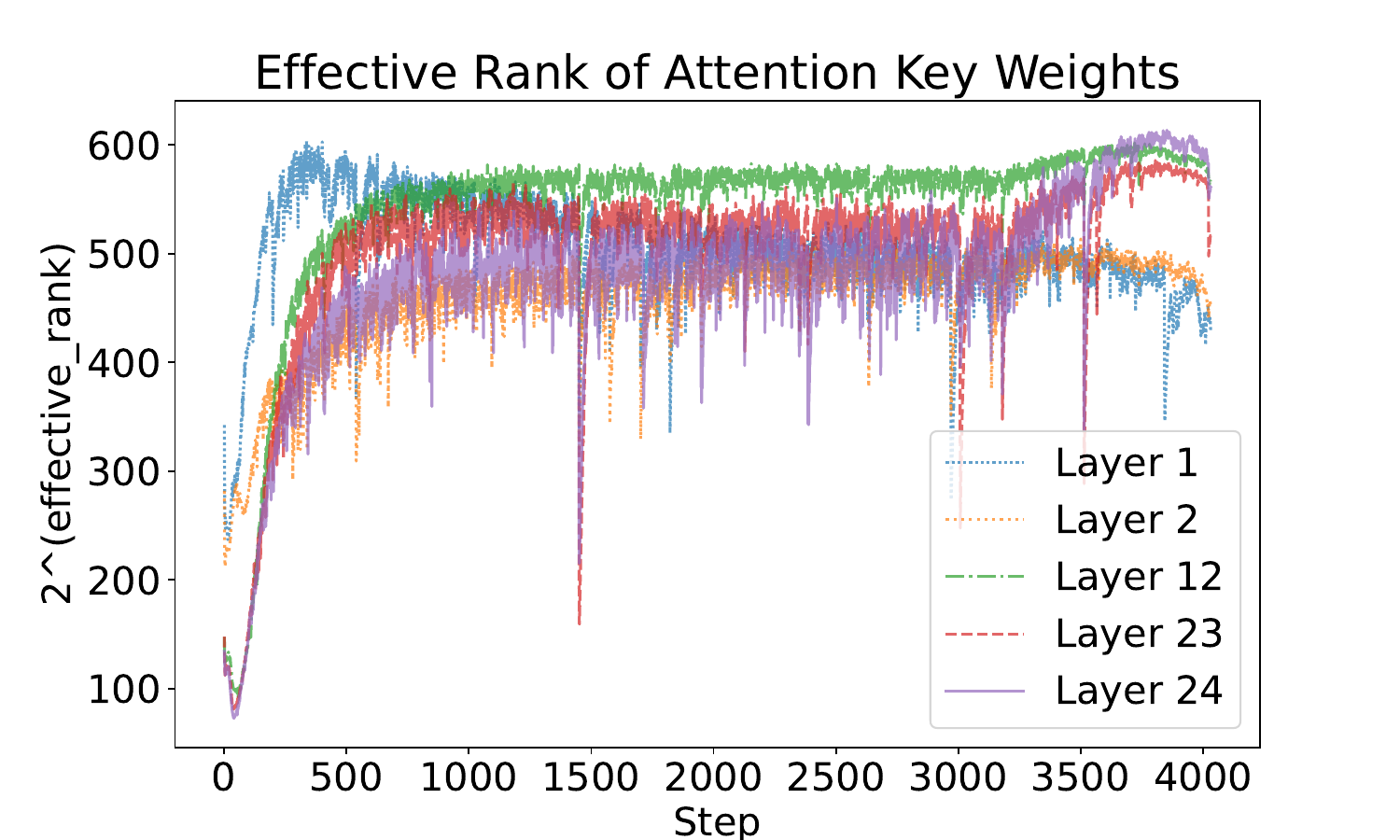}
        \caption{}
        \label{attn_k}
    \end{subfigure}
    \hfill
    \begin{subfigure}[b]{0.45\textwidth}
        \centering
        \includegraphics[width=\textwidth]{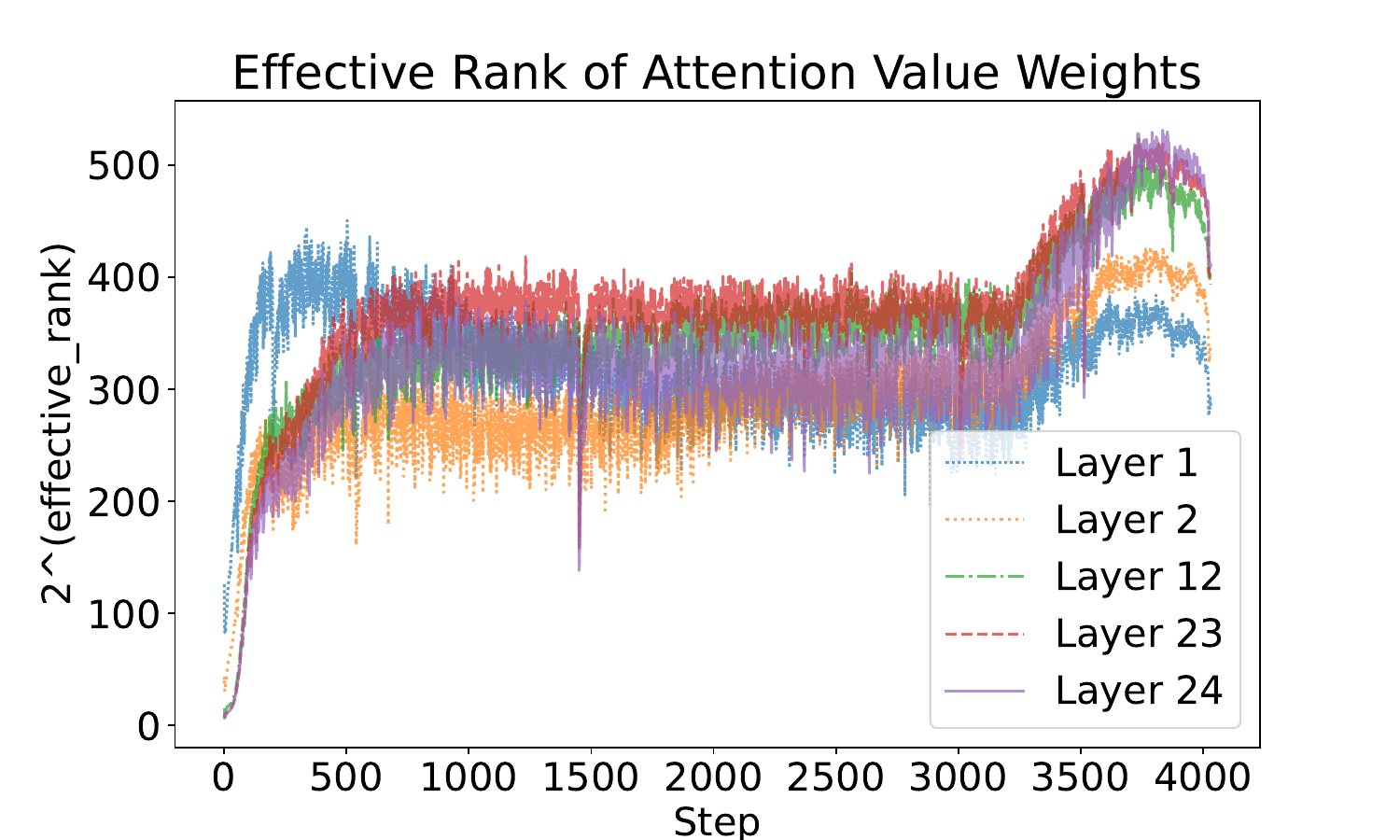}
        \caption{}
        \label{attn_v}
    \end{subfigure}
    
    \caption{$2^{\text{erank}(M)}$ during training of the 400M model from different layers for the projection and fully connected weights in MLP, the projection and QKV weights in attention layers respectively. The full rank for reference is 1024. }
    \label{fig:effective_rank_400M}
\end{figure}

In \autoref{fig:effective_rank_400M}, we show the approximation of $2^{\text{erank}(M)}$ during training for the 400M model, following the default training configuration in \autoref{sec:hyper} and model architecture in \autoref{tab:scale_cfg}. We observe that:
\begin{itemize}
    \item  In most cases, the middle layers (like layer 12) have relatively higher effective rank than the early (like layer 1) or near-end layers (like layer 24). Also, the rest of the layers have very similar values and trends like layer 12 so we omit them in the plot.
    \item The effective rank usually increases very quickly during the first few optimization steps, with a potential sudden rise in the learning rate decay period. This also corresponds to a steeper loss curve, meaning that the better loss decrease comes from more actively explored directions in the update. At the end of learning rate decay where the learning rate is very close to 0, the effective rank also decreases. 
    \item Most of the values are in the $[400, 800]$ range.
\end{itemize}

\begin{figure}
    \centering
    \includegraphics[width=0.6\linewidth]{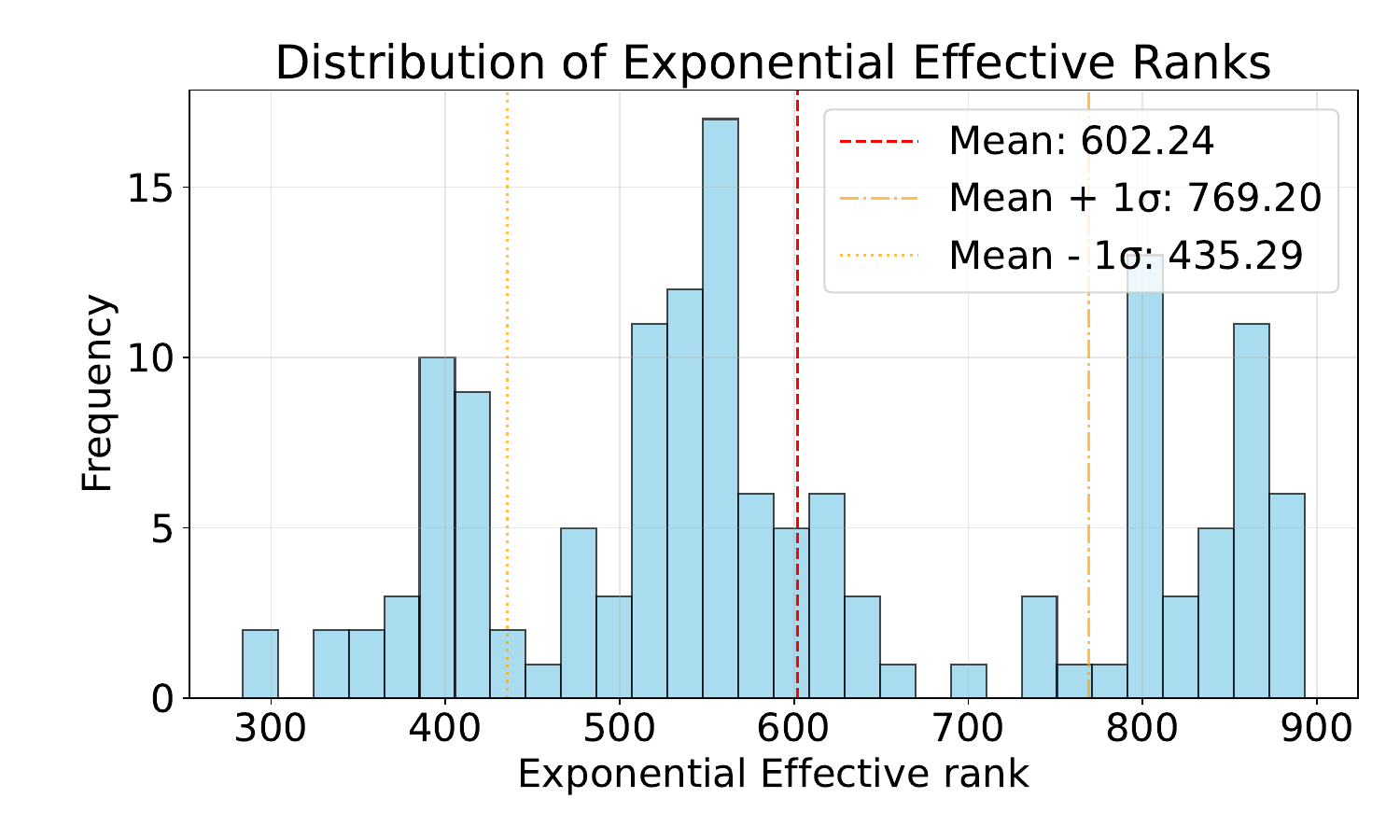}
    \caption{Effective rank distribution from the 400M model.}
    \label{fig:effective_rank}
\end{figure}

We also visualize the distribution of the approximated $2^{\text{erank}(M)}$ value at the end of training in \autoref{fig:effective_rank}. There are approximately three peak values around rank 400, 600 and 800, which corresponds to the value weights in attention, the key/query weights in attention, and the MLP weights in \autoref{fig:effective_rank}. 

\textbf{Filtering the column of $W$, and adaptive rank selection.} With the observations above, a natural question is: ``\emph{Could the effective rank guide us to perform adaptive rank selection in Dion?}'' We consider several adaptive techniques: 1) Filtering: select a threshold of the smallest column norm of $W$ and ignore all columns with smaller norms; 2) Effective rank: recompute the QR with the rank equivalent to the precomputed exponential effective rank.

\begin{table}[h]
\centering
\begin{tabular}{llll}
\toprule
Full rank                       & Filtering                       & Effective Rank                   & 75\% Rank                       \\
\hline
{3.003±0.001} & {\textbf{3.002±0.001}} & 
{3.013±0.005} & {3.015±0.004}\\
\bottomrule
\end{tabular}
\vspace{1mm}
\caption{The final validation loss from the training of the 400M model using different rank configurations. ``Filtering'' indicates filtering the columns of $W$ with norms smaller than $5e-6$. The results are averaged across 3 runs using different random seeds.}
\label{tab:adaptive_rank}
\end{table}

In \autoref{tab:adaptive_rank}, we compare the result from training with full rank, training with rank filtering, training with exponential effective rank, and training with 75\% rank\footnote{Notice that 75\% rank is also covering most of the effective rank we show in \autoref{fig:effective_rank}. 
}. For filtering, we tried the threshold values $\{1e-6, 5e-6, 1e-5\}$ and found a better performance with $5e-6$ which is shown in \autoref{tab:adaptive_rank}. For effective rank, we warm up the model training with full rank for the first 200 steps and then switch to effective rank afterwards. We observe that full rank and filtering have very similar final performance, and filtering can even achieve slightly better performance; using effective rank and 0.75 rank could be slightly worse with a small 0.01 loss gap on the validation performance. 

Although the proposed methods cannot directly improve the speed of training due to the fact that filtering and recomputing QR are both done after the normal QR computation, the results still showcase a promising future direction that utilizes adaptive rank selection to improve the update quality/speed trade-off.

\section{Detailed Analysis of Complexities}

  \subsection{Detailed Computational Complexity of Unsharded Dion} 
\label{details:comp}

We provide a more detailed analysis of the number of FLOPs needed by Dion versus Muon for optimizing a single $m \times n$ parameter matrix. For simplicity, we omit all element-wise operations (e.g.~updating momentum with gradient), as they do not affect the asymptotic runtime.

The following FLOP counts are assumed:
\begin{itemize}
    \item \textbf{Matrix multiplication:} Multiplying $m \times n$ and $n \times p$ matrices requires $2mnp$ FLOPs.
    \item \textbf{QR decomposition:} For a $m \times n$ matrix with $m \geq n$, the Householder QR algorithm requires $2mn^2 - (2/3)n^3$ FLOPs \citep{big6matrixfactorizations}.
    \item \textbf{Cholesky decomposition:} For a $n \times n$ square matrix, Cholesky decomposition requires $n^3 / 3$ FLOPs \citep{big6matrixfactorizations}.
    \item \textbf{Solve triangular:} Solving a linear system involving a $n \times n$ triangular matrix and a vector, using forward- or back-substitution, requires $n^2$ FLOPs. Therefore, computing $AR^{-1}$ for a $m \times n$ matrix $A$ and $n \times n$ triangular matrix $R$ requires $m$ matrix-vector solves, for a total of $mn^2$ FLOPs.
\end{itemize}

For one update step of Unsharded Dion (\autoref{alg:un-dion}), the major operations and costs are:
\[
\renewcommand{\arraystretch}{1.3}
\begin{array}{c|c|c}
    \textbf{Operation} & \textbf{Shape} & \textbf{FLOPs} \\ \hline
    M V & (m\times n)(n\times r) & 2mnr \\
    M^\top U & (n\times m)(m\times r) & 2mnr \\
    S P & (1.25r \times m)(m\times r) & 2.5mr^2 \\
    \texttt{QR}(\tilde P) & (1.25r \times r) & \tfrac{11}{6}r^3 \\
    P R_1^{-1} \text{ (triangular solve)} & (m\times r)(r\times r) & mr^2 \\
    B^\top B & (r\times m)(m\times r) & 2mr^2 \\
    \texttt{Cholesky}(G) & (r\times r) & \tfrac{1}{3}r^3 \\
    B R_2^{-1} \text{ (triangular solve)}& (m\times r)(r\times r) & mr^2 \\
    U W\top & (m\times r)(r\times n) & 2mnr\\
    U V^\top & (m\times r)(r\times n) & 2mnr
\end{array}
\]

Summing these contributions yields
\[
8mnr \;+\; 6.5\,mr^2 \;+\; 2.17\,r^3 .
\]

For Muon, each Newton–Schulz iteration requires three matrix multiplications. For $m \geq n$, this is
\[
(2mn^2 + 2n^3) + (2mn^2) = 4mn^2 + 2n^3.
\]
With the default five iterations, the total cost is
$20mn^2 + 10n^3$.

 \subsection{Detailed Communication Volumes for 2D-Sharded Dion}
\label{app:comm-details}
 
Let $m=|I|$, $n=|J|$, $r=|R|$. Mesh sizes are $a=|\texttt{X}|$ and $b=|\texttt{Y}|$.
Batch axis $L$ has $|L|=b$ in the \emph{small-batch} regime and $|L|=ab$ in the \emph{large-batch} regime.
DP synchronization along \texttt{Z} is omitted (see \autoref{sec:compressed-sync}).
Payloads are reported in \emph{elements}.

\textbf{Cost model (bidirectional ring).}
We assume a 1D \emph{bidirectional} ring, i.e., both directions are used concurrently as in modern ICI interconnects~\citep[Ch.~3]{scaling-book}.
Let a collective operate on $V$ elements per device and let $W_{\mathrm{ici}}^{\leftrightarrow}$ denote  the bidirectional ICI per-device bandwidth.
Ignoring latency, the bandwidth-bound times are
\begin{align*}
T_{\mathrm{RS}}(V) &= \frac{V}{W_{\mathrm{ici}}^{\leftrightarrow}}, &
T_{\mathrm{AG}}(V) &= \frac{V}{W_{\mathrm{ici}}^{\leftrightarrow}}, \\
T_{\mathrm{AR}}(V) &= \frac{2V}{W_{\mathrm{ici}}^{\leftrightarrow}}, &
T_{\mathrm{A2A}}(V) &\approx \frac{V}{4\,W_{\mathrm{ici}}^{\leftrightarrow}}.
\end{align*}
Here, AllReduce is modeled as ReduceScatter followed by AllGather; in a 1D bidirectional ring, AllToAll costs about one quarter of an AllGather.
If latency is modeled, a ring requires $p/2$ pipeline stages (versus $p-1$ for a unidirectional ring), contributing an additive term $\tfrac{p}{2}t_{\mathrm{hop}}$, but we ignore this term in the volumes and times below.
 
\begin{center}
\small
\setlength{\tabcolsep}{4pt}
\renewcommand{\arraystretch}{1.2}
\begin{tabular}{@{}llll@{}}
\toprule
\textbf{Axis/Regime} & \textbf{Collective} & \textbf{Payload $V$} & \textbf{Time $T$} \\
\midrule
\multicolumn{4}{@{}l}{\textbf{Across \texttt{X}} (col-sharded, $J_{\texttt{X}}$)}\\
SB ($|L|{=}b$)  & $\AllReduce(P)$ & $mr$ & $\tfrac{2mr}{W_{\mathrm{ici}}^{\leftrightarrow}}$ \\
LB ($|L|{=}ab$) & $\ReduceScatter(P){+}\AllGather(U)$ & $a m r$ each & $\tfrac{2 a m r}{W_{\mathrm{ici}}^{\leftrightarrow}}$ \\
Both            & $\AllReduce(\mathrm{col\ norms})$ & $|L|r$ (SB:$br$, LB:$abr$) & $\tfrac{2|L|r}{W_{\mathrm{ici}}^{\leftrightarrow}}$ \\
\midrule
\multicolumn{4}{@{}l}{\textbf{Across \texttt{Y}} (row-sharded, $I_{\texttt{Y}}$)}\\
SB ($|L|{=}b$)  & $\AllReduce(W)$ & $\tfrac{b}{a}nr$ & $\tfrac{2 b n r/a}{W_{\mathrm{ici}}^{\leftrightarrow}}$ \\
LB ($|L|{=}ab$) & $\AllReduce(W)$ & $bnr$ & $\tfrac{2 b n r}{W_{\mathrm{ici}}^{\leftrightarrow}}$ \\
Ortho.          & $\ReduceScatter(S\!\cdot P)$ & $1.25 b r^2$ & $\tfrac{1.25 b r^2}{W_{\mathrm{ici}}^{\leftrightarrow}}$ \\
Ortho.          & $\AllGather(R_1)$ & $b r^2$ & $\tfrac{b r^2}{W_{\mathrm{ici}}^{\leftrightarrow}}$ \\
Ortho.          & $\ReduceScatter(G)$ & $b r^2$ & $\tfrac{b r^2}{W_{\mathrm{ici}}^{\leftrightarrow}}$ \\
Ortho.          & $\AllGather(R_2)$ & $b r^2$ & $\tfrac{b r^2}{W_{\mathrm{ici}}^{\leftrightarrow}}$ \\
\bottomrule
\end{tabular}  
\end{center}
The main takeaways from the analysis are summarized below:
\begin{itemize}[leftmargin=1.5em]
\item Along \texttt{X}, the dominant traffic is $O(mr)$ from the $P/U$ exchange, with an additional $O(|L|r)$ from column-norm reductions.
\item Along \texttt{Y}, communication is $O(nr)$ from the $W$ reduction, plus only $O(r^2)$ from the orthonormalization micro-collectives.
\item Importantly, no collective scales with $mn$. The overall communication is linear in $mr$ or $nr$, with only quadratic $r^2$ terms. This scaling is highly favorable when $r\ll m,n$.
\end{itemize}

\section{Update Rules for Non-Matrix Parameters}
\label{sec:non-matrix}

We now detail the use of Adam and Lion for scalar parameters, and explain the normalization and scaling rules summarized in \autoref{tab:scaling-factors}. The goal is to ensure consistent learning rate transfer across model sizes—\emph{one base learning rate, two optimizers, all parameter types}.

\subsection{Normalization and Scaling for Learning Rate Transfer}

Following \citet{yang2023spectral}, effective transfer requires that both parameters and their updates maintain a $\Theta(1)$ \emph{natural norm}: the RMS norm\footnote{For a vector $\mathbf{v} \in \R^d$, the RMS norm is $\|\mathbf{v}\|_{\text{RMS}} := \frac{1}{\sqrt{d}} \|\mathbf{v}\|_2$.} for dense vectors  (e.g., activations), the $\ell_2$ norm for sparse vectors (e.g., one-hot encodings), and the operator norm for matrices.

Dion satisfies this condition through orthonormalized matrix updates, while Adam produces an approximately constant RMS update for scalars, and Lion guarantees a constant RMS update. Starting from a single base learning rate, we apply dimension-dependent scale factors so that all parameter types achieve unit natural norm. Key cases include:
\begin{itemize}
    \item \textbf{Weight matrix:} For $d_{\text{out}} \times d_{\text{in}}$ weights, scaling by $\sqrt{d_{\text{out}}/d_{\text{in}}}$ ensures a unit RMS operator norm \citep{bernstein2025deriving}.
    \item \textbf{Bias \& embeddings:} Normalize updates to unit RMS norm.
    \item \textbf{LM Head:} Scale unit RMS updates by $1/\sqrt{d_{\text{in}}}$ (see \autoref{sec:unembedding-scale-factor}).
    \item \textbf{Normalization parameters:} Treated as $1 \times 1$ matrices or vectors; all scale factors are $1$.
\end{itemize}

\begin{figure}[t]
\centering
\begin{subfigure}{0.49\linewidth}
\centering
\includegraphics[width=\linewidth]{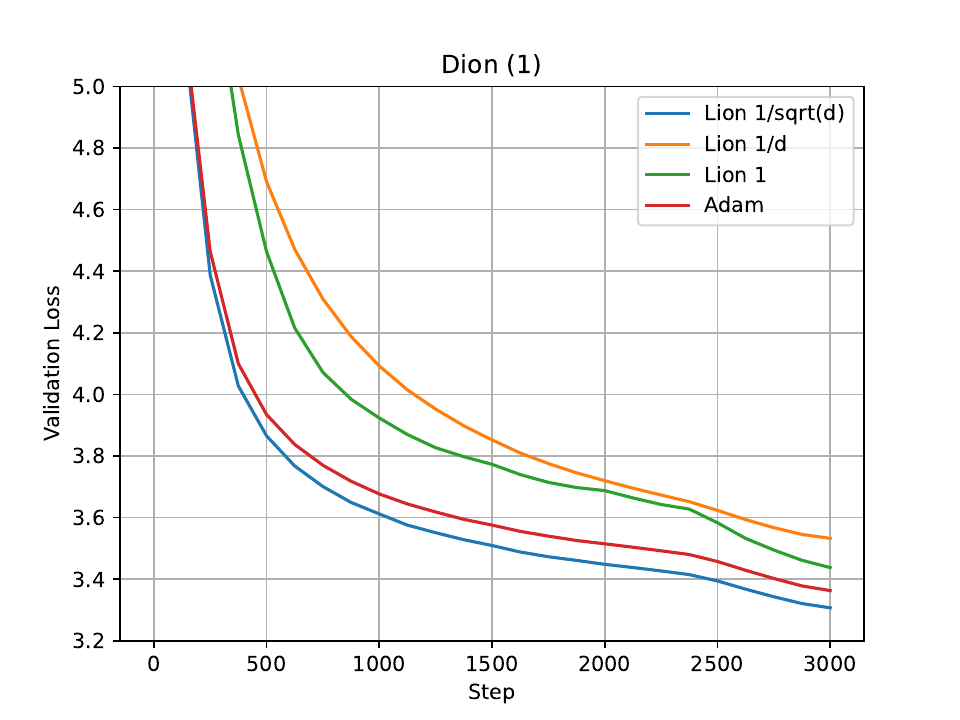}
\end{subfigure}
\begin{subfigure}{0.49\linewidth}
\centering
\includegraphics[width=\linewidth]{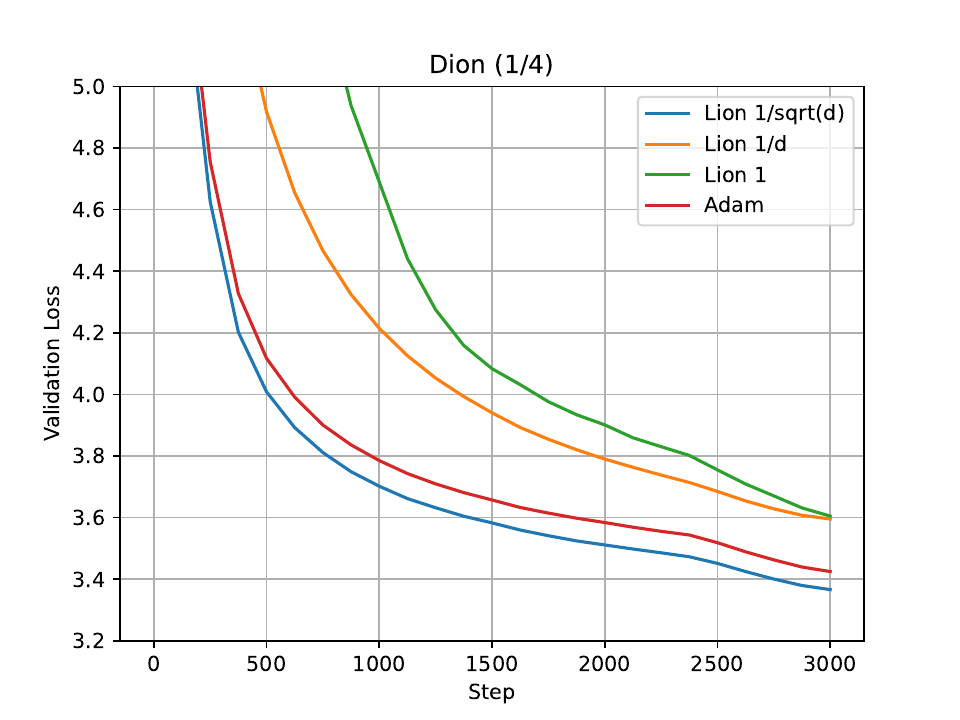}
\end{subfigure}

\vspace*{-10pt}
\begin{subfigure}{0.49\linewidth}
\centering
\includegraphics[width=\linewidth]{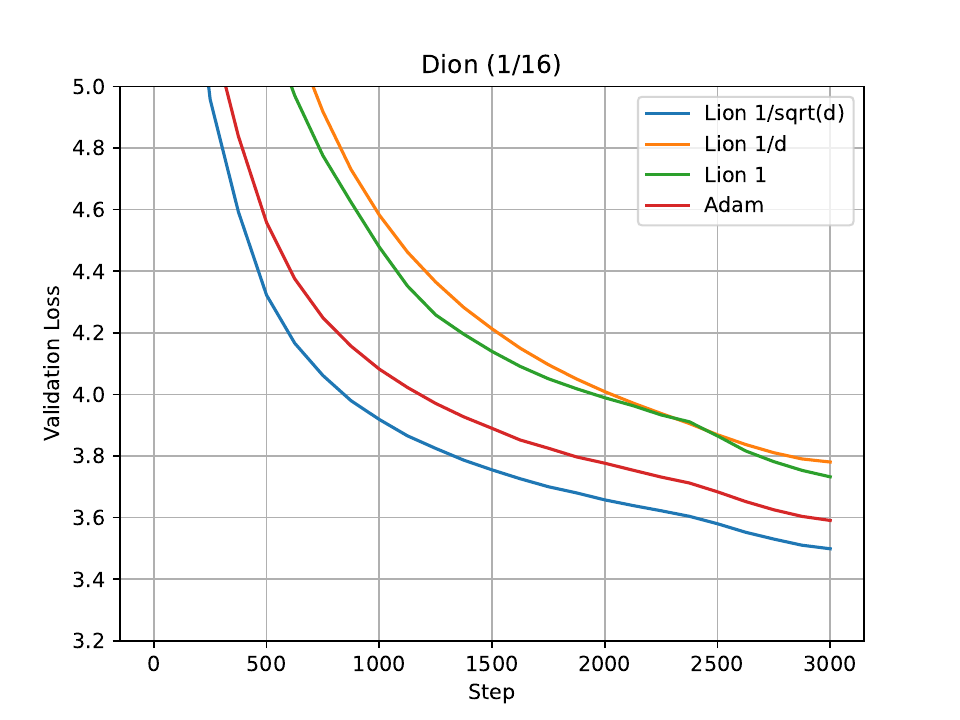}
\end{subfigure}
\begin{subfigure}{0.49\linewidth}
\centering
\includegraphics[width=\linewidth]{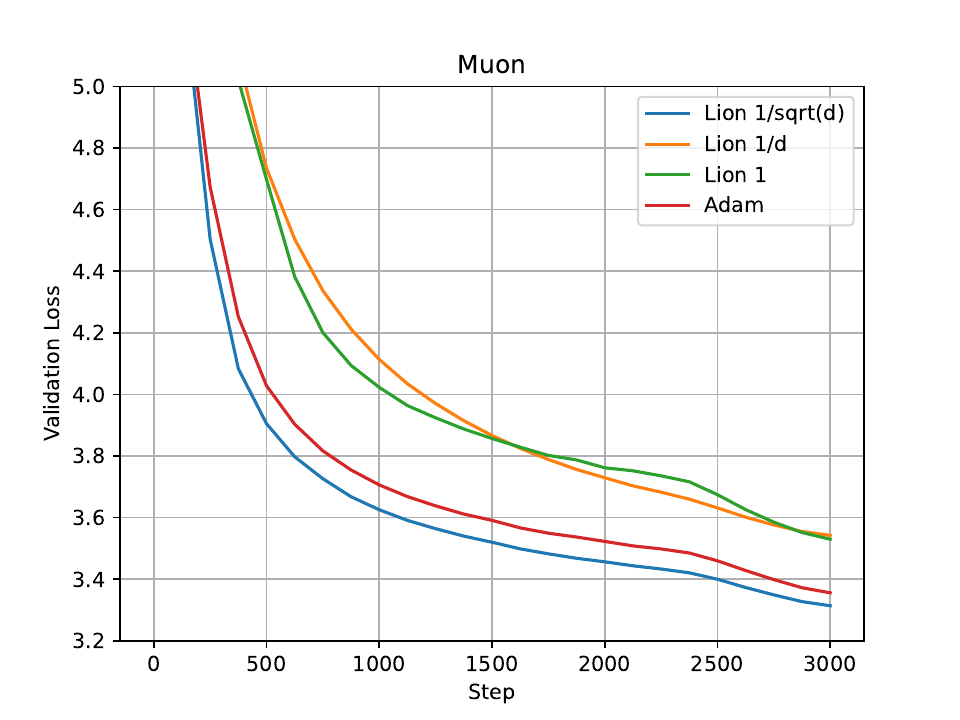}
\end{subfigure}

\caption{\footnotesize 
Comparison of optimizers and learning rate scale factors for the LM head. 
Lion is tested with learning rates scaled by $1/\sqrt{d}$, $1/d$, or left unscaled, 
while Adam uses a single fixed rate for all non-matrix parameters. 
Scaling by $1/\sqrt{d}$ with Lion consistently yields the best results, 
across Dion rank fractions ($1$, $1/4$, $1/16$) and when applied to Muon.}
\label{fig:unembedding-scale}
\end{figure}
\subsection{Scaling Factor for LM Head Parameters}
\label{sec:unembedding-scale-factor}

For an unembedding vector $\mathbf{v}$ and activation $\mathbf{h} \in \R^d$, the logit is $s = \mathbf{v} \cdot \mathbf{h}$. If updates $\Delta \mathbf{v}$ are i.i.d.\ with zero mean,
\[
\E[(\Delta s)^2] = d \cdot \|\Delta \mathbf{v}\|_\text{RMS}^2 \|\mathbf{h}\|_\text{RMS}^2.
\]
Assuming $\|\mathbf{h}\|_\text{RMS} = 1$, we have $\E[\|\Delta s\|_\text{RMS}] = \sqrt{d}\,\|\Delta \mathbf{v}\|_\text{RMS}$. To achieve $\Theta(1)$ change in logits, we normalize to $\|\Delta \mathbf{v}\|_\text{RMS} = 1$ and scale by $1/\sqrt{d}$. This differs from the $1/d$ factor of \citet{yang2023spectral}, which ensures a worst-case bound but is overly conservative. Empirically, $1/\sqrt{d}$ yields stable training and the best validation loss.

\autoref{fig:unembedding-scale} compares three scale factors ($1/\sqrt{d}$, $1/d$, $1$) using Lion as the scalar optimizer. $1/\sqrt{d}$ consistently outperforms both alternatives and even beats a manually tuned Adam baseline, while the unscaled case ($1$) causes gradient explosions and instability.

\subsection{Scalar Optimizers: Adam vs.\ Lion}

Adam \citep{kingma2014adam} and AdamW \citep{loshchilov2018decoupled} are widely used for scalar parameters. However, Adam requires a separately tuned learning rate, breaking the goal of a single transferable schedule. By contrast, Lion \citep{chen2023symbolic}, which uses sign-based updates, guarantees a constant RMS norm of one. Combined with the scale factors in \autoref{tab:scaling-factors}, this allows Dion and Lion to share a single base learning rate across all parameters. Empirically, Dion+Lion outperforms Dion+Adam (see \autoref{fig:unembedding-scale}) and requires no extra tuning.

These results highlight the shortcomings of the common practice of using one Adam learning rate for all parameters. Embedding and unembedding layers alone differ in ideal update magnitudes by a factor of $\sqrt{d_\text{model}}$, often exceeding $100$ in large models. A single Adam learning rate must compromise between them, slowing convergence. Weight-tying the embedding and unembedding layers may partially mask this mismatch but is likely detrimental overall.

\section{Hyperparameter Choices}
\label{sec:hyper}

Our default model configurations are given in \autoref{tab:scale_cfg}. Unless otherwise specified, our experiments use the 160M parameter model size. All models use the GPT2 tokenizer with a vocabulary size of 50304. We use rotary position embeddings \citep{su2023rotary}, non-parametric RMSNorm, and omit biases from linear layers. The activation function for MLP layers is squared ReLU \citep{so2022primer}.

We train all models on the FineWeb \citep{penedo2024the} or FineWeb-Edu \citep{lozhkov2024fineweb-edu} datasets. Unless otherwise specified, we train on an approximately Chinchilla-optimal number of tokens \citep{hoffmann2022training} (tokens $\approx 20\times$ model parameters) for each model size. We use NVIDIA H100 or AMD MI300X GPUs for all experiments.

\begin{table}[h]
    \centering\small
    \begin{tabular}{@{}cccccc@{}}
        \toprule
        Model & $d_\text{model}$ / Layers / Heads & Batch Size & Total Steps & Total Tokens  \\
        \midrule
        160~M &  768 / 12 /  6 & 1.0~M & 3~K  & 3.1 B \\
        350~M & 1024 / 24 / 32 & 2.1~M & 4~K  & 8.4 B \\
        1.3~B & 2048 / 24 / 32 & 2.1~M & 12~K & 25.2 B \\
        3~B & 3072 / 24 / 32 & 2.1~M & 30~K & 63 B  \\
        \bottomrule
    \end{tabular}
    \vspace{4pt}
    \caption{\footnotesize Default configurations for each model size.}
    \label{tab:scale_cfg}
\end{table}




\subsection{Results in \autoref{sec:model}}
\label{details:model}

 

We use the model configurations detailed in \autoref{tab:scale_cfg}.
For both Muon and Dion, we use a fixed learning rate of $0.01$ across all model sizes. This choice is supported by our hyperparameter transfer results (\autoref{fig:transfer}), which show that the optimal learning rate remains stable across scale.

\subsection{Results in \autoref{sec:transfer}}
\label{details:transfer}

Each model size is trained with its respective Chinchilla-optimal number of tokens on the FineWeb-Edu dataset. All runs use a bath size of 1M tokens.  We use a constant learning rate schedule with no warmup and $10\%$ linear cooldown. We use Adam with learning rate $0.002$ and $(\beta_1, \beta_2) = (0.9, 0.95)$ as the optimizer for non-matrix parameters.

\end{document}